\documentclass[10pt,a4paper,twocolumn]{article}

\usepackage[utf8]{inputenc}
\usepackage[T1]{fontenc}
\usepackage{amsmath,amssymb,amsfonts}
\usepackage{graphicx}
\usepackage[dvipsnames]{xcolor}
\usepackage[english]{babel}
\usepackage{geometry}
\usepackage{microtype}
\usepackage{listings}
\usepackage{caption}
\usepackage{subcaption}
\usepackage{booktabs}
\usepackage{longtable}
\usepackage{multirow}
\usepackage{array}
\usepackage{float}
\usepackage{textcomp}
\usepackage{balance}
\usepackage{authblk}
\usepackage[authoryear,round]{natbib}

\usepackage{hyperref}
\hypersetup{
    colorlinks=true,
    linkcolor=MidnightBlue,
    citecolor=MidnightBlue,
    urlcolor=MidnightBlue
}

\newcommand{\textgreaterequal}{\ensuremath{\geq}}

\title{From Photons to Physics: Autonomous Indoor Drones and the Future of Objective Property Assessment}

\author[1,*]{Petteri Teikari}
\author[1,2]{Mike Jarrell}
\author[1]{Irene Bandera Moreno}
\author[1]{Harri Pesola}

\affil[1]{Mill Hill Garage}
\affil[2]{JB Real Estate Valuation \& Advisory, LLC}
\affil[*]{Corresponding author: petteri@millhillgarage.com}

\begin{document}

\maketitle

\begin{abstract}
The convergence of autonomous indoor drones with physics-aware sensing technologies promises to transform property assessment from subjective visual inspection to objective, quantitative measurement. This comprehensive review examines the technical foundations enabling this paradigm shift across four critical domains: (1) platform architectures optimized for indoor navigation, where weight constraints drive innovations in heterogeneous computing, collision-tolerant design, and hierarchical control systems; (2) advanced sensing modalities that extend perception beyond human vision, including hyperspectral imaging for material identification, polarimetric sensing for surface characterization, and computational imaging with metaphotonics enabling radical miniaturization; (3) intelligent autonomy through active reconstruction algorithms, where drones equipped with 3D Gaussian Splatting make strategic decisions about viewpoint selection to maximize information gain within battery constraints; and (4) integration pathways with existing property workflows, including Building Information Modeling (BIM) systems and industry standards like Uniform Appraisal Dataset (UAD) 3.6.

---
\end{abstract}

\section{Introduction}

The property technology sector faces a fundamental measurement challenge. The U.S. real estate appraisal market, valued at \$11.9 billion in 2023, relies on critical decisions about value, risk, and condition that depend primarily on subjective human judgment and visual inspection methods. This disconnect between economic importance and measurement precision manifests in misvalued properties, undetected defects, disputed insurance claims, and systematic lending risks. While industries from manufacturing to medicine have adopted objective, sensor-based measurement, property assessment remains dependent on human perception---creating what we term the ``Objectivity Gap'': the disparity between a property's true physical state and its assessed condition based on limited visual observation.

Contemporary property evaluation methodologies exhibit systematic subjectivity that compromises market efficiency. Persistent size error is the first pillar of the objectivity gap. Recognising that valuation hinges on accurate area, Government-Sponsored Enterprises (GSEs) made the ANSI\,Z765‑2021 measurement standard mandatory for all conforming loans, Fannie Mae 1\,April\,2022 and Freddie Mac November 2, 2023, expressly to ``establish a consistent, repeatable process'' for dwelling‑size calculations \citep{institute2025}. Yet the requirement highlights the problem: traditional tape‑measure practice still produces materially different numbers---and therefore different values---for the same property.

These inconsistencies are exacerbated by a shrinking talent pipeline: 62\,percent of U.S. appraisers are now over 51\,years old, while only 13\,percent are 35 or younger, limiting the industry's capacity to train new practitioners and elongating report turn‑times (﻿\citep{realtor2025}). Together, measurement variance, rating drift, and an aging workforce underscore the need for objective, sensor‑based methods that can deliver consistent property data independent of individual expertise.

Beyond valuation, latent defects represent a critical information failure. Standard property policies exclude inherent defects, creating uninsurable risks that manifest catastrophically, as demonstrated by Canada's building envelope crisis \citep{blc2018}. The emergence of specialized Latent Defect Insurance markets \citep{marsh2023, lockton2020} signals the magnitude of this systematic measurement failure.

Current property inspection relies fundamentally on what the human eye can perceive. An appraiser walks through a home, manually measures dimensions, photographs surfaces, and assigns subjective condition ratings that can affect property values substantially. This visual-centric approach faces inherent limitations: RGB (color, red-green-blue) cameras cannot distinguish between genuine hardwood and high-quality laminate, standard photography cannot detect water intrusion behind walls, and subjective condition ratings mandated by standards like Uniform Appraisal Dataset (UAD) 3.6 remain inconsistent between field professionals. The solution lies not in incremental improvements to visual inspection but in extending sensing capabilities beyond human perception---measuring what materials \emph{are} rather than how they \emph{appear}. Advanced sensing modalities like hyperspectral and polarimetric imaging can reveal material composition and physical condition through spectral signatures and polarization states.

The deployment of autonomous drones equipped with physics-aware sensors offers a systematic approach to property assessment. Collections of lightweight platforms could map buildings while creating comprehensive physical digital twins encoding material composition and condition. However, the path from laboratory demonstrations to commercial deployment faces significant technical challenges. Indoor drones face stringent size, weight, and power (SWaP) constraints---typical drones suitable for indoor operations such as ModalAI Starling 2 Max allow 500g payload, leaving minimal allocation for advanced sensors after essential flight systems. Current hyperspectral sensors exceed these weight budgets, real-time multimodal sensor fusion surpasses available edge computing capabilities, and metasurface-based computational imaging solutions remain at Technology Readiness Levels (TRL) 3-4. The fundamental SWaP constraint requires choosing between flight time, sensor payload, and computational power.

Beyond technical challenges, market adoption faces substantial barriers. The appraisal industry, built on established practices and regulatory frameworks, shows limited enthusiasm for technologies that could automate portions of their work. While physics-based inspection could provide material authentication and moisture detection capabilities, willingness to pay remains unproven---property buyers already hesitate at traditional inspection costs. Insurance companies express interest in pre-loss baselines but remain uncommitted to funding widespread deployment. The classic technology adoption challenge emerges: without market demand, investment in miniaturization stalls; without miniaturized solutions, market adoption remains limited.

Near-term deployment will likely follow a phased approach: first, conventional sensors with intelligent exploration; then gradually incorporating advanced modalities as miniaturization progresses. Critical challenges include developing material spectral databases, standardizing data formats for industry integration, and validating measurement accuracy across diverse indoor conditions. While these technologies promise objective property assessment, they primarily address measurement accuracy rather than systemic valuation biases. Success requires coordinated advancement in sensor miniaturization, edge computing, and industry workflow integration. This review provides the first comprehensive analysis of physics-aware sensing specifically for autonomous indoor property inspection, synthesizing advances across robotics, optics, and computer vision while maintaining critical perspective on technological readiness and practical limitations.

This review argues that autonomous indoor drones equipped with physics-aware sensing technologies represent a critical pathway toward objective property measurement, though their deployment faces significant technical hurdles in sensor miniaturization, computational requirements, and system integration that will necessitate a phased implementation approach over the coming decade.

\section{The Autonomous Indoor Inspection Platform}

While most Unmanned Aerial Vehicle (UAVs, i.e. drones) face severe size, weight, and power (SWaP) constraints regardless of operating environment (\citep{tomic2012, hassanalian2017, hadidi2021}), indoor drone design presents fundamentally different engineering challenges than outdoor counterparts. The constraints of confined spaces, stringent weight regulations, and requirements for millimeter-precision sensing create a unique design paradigm demanding comprehensive reconsideration of platform architecture---from aerodynamics to artificial intelligence.

\subsection{The Indoor Design Challenge: Why One Size Fails All}

The notion of a universal drone design proves impractical given the radically different operational environments. Indoor inspection drones operate under fundamentally different constraints than their outdoor counterparts, leading to divergent design philosophies and engineering tradeoffs (\citep{hadidi2021, krishnan2021, dibs2024, mourtzis2024, patel2024, dominguez2025}). Outdoor construction drones exemplified by the DJI Matrice 350 RTK prioritize substantial payload capacity (\citep{shelare2024, amale2024, sourirajan2025}) to support industrial sensors, extended flight duration (55 minutes) for comprehensive site coverage, and robust wind resistance (15 m/s) enabling all-weather operation. Military platforms \citep{karali2024} such as the Skydio X2D optimize for thermal imaging capabilities, encrypted communications, and survivability features essential for tactical operations.


In stark contrast, indoor inspection drones must optimize for entirely different parameters: ultra-lightweight construction (\textless500g total weight) to comply with regulatory frameworks, exceptional hover efficiency (\citep{hu2021, yoon2023, yoon2025}), collision tolerance rather than avoidance given inevitable surface proximity, and sophisticated sensor fusion algorithms for GPS-denied navigation. Additionally, these platforms must maintain minimal acoustic signatures to enable operation in occupied buildings without disruption (\citep{miljkovi2018, bian2021, kawai2024, macke2024, lotinga2025}).

The ModalAI Starling 2 Max exemplifies this indoor-specific design philosophy. With a 500g airframe weight and 500g payload capacity, it deliberately sacrifices the raw lifting power of outdoor platforms to achieve the agility and precision required for navigation in residential and commercial interiors. This represents not a limitation but a deliberate optimization for environments where every gram affects maneuverability through tight spaces and every decibel impacts acceptability in occupied spaces \citep{hadidi2021}.

\begin{figure}
    \centering
    \includegraphics[width=1\linewidth]{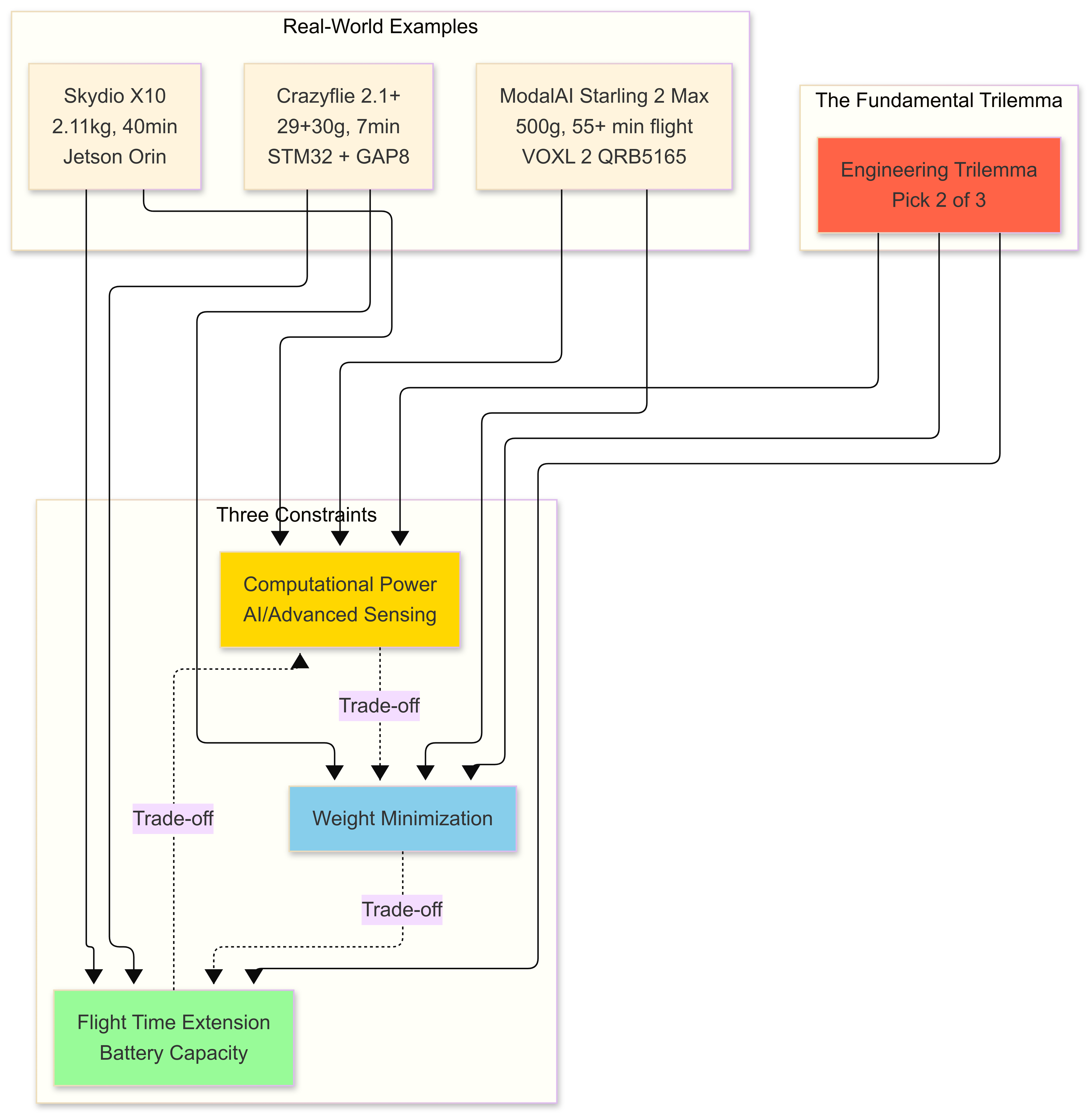}
    \caption{Engineering trilemma: computational power, weight, and flight time cannot be simultaneously maximized. ModalAI Starling 2 Max (500g, 55-minute, custom SoC) balances computation with weight compliance. Skydio X10 (2.1kg, 40-minute, Jetson Orin) prioritizes computing/endurance but exceeds indoor limits. Crazyflie 2.1+ (29g, 7-minute, STM32 MCU + 30g GAP8 accelerator) achieves miniaturization but lacks computational resources. Bidirectional constraints reflect fundamental physics limitations of battery density, processor efficiency, and aerodynamics.}
    \label{fig:trilemma}
\end{figure}

Drone design confronts an engineering trilemma as unforgiving as fundamental physics (\autoref{fig:trilemma}): simultaneous maximization of computational power \citep{wanIntelligenceRoboticComputing2023}, weight minimization \citep{palomba2022, kowalik2025}, and flight time extension (\citep{akram2025, ahmed2025, anoune2025}) proves impossible. This constraint shapes every design decision and necessitates fundamental architectural innovations rather than incremental improvements.

\subsection{Heterogeneous Computing: The Architecture of Aerial Intelligence}

The computational demands of autonomous indoor inspection necessitate a hierarchical, heterogeneous architecture that strategically distributes processing across specialized hardware components (\citep{ahmed2022, mika2023, liu2024}). Modern drone platforms orchestrate multiple computational paradigms, each optimized for specific tasks within the perception-action pipeline, creating a sophisticated computational ecosystem \citep{chen2024, liu2025}.

At the foundational level, microcontrollers (MCUs) provide deterministic real-time control essential for flight stabilization and safety-critical functions. While computationally limited compared to application processors, recent advances in TinyML \citep{cereda2024} and Federated Learning \cite{dasilva2025} demonstrate that even resource-constrained MCUs can participate meaningfully in distributed learning paradigms. These capabilities enable drones to train local models and share learned parameters rather than raw data, facilitating collective intelligence while respecting bandwidth and privacy constraints.

For tasks requiring hardware-level parallelism and microsecond-level latency, Field-Programmable Gate Arrays (FPGAs) excel at accelerating specific computational workloads, particularly those involving dense fixed-point operations or irregular parallelism patterns \citep{wang2025, nieto2025}. The Xilinx Zynq UltraScale+ MPSoC exemplifies modern heterogeneous computing by combining FPGA fabric with ARM processors on a single die \citep{serranocases2024}, making it particularly suitable for sensor fusion algorithms and deterministic control loops that feed into higher-level planning systems.

Dedicated AI accelerators \citep{garreau2025} represent another crucial component of the computational hierarchy. The Hailo-8 M.2 Module delivers exceptional neural network inference performance per watt through specialized hardware architectures optimized for convolution and matrix operations. Similarly, the GreenWaves Technologies GAP8, a parallel ultra-low-power processor specifically designed for AI applications, powers platforms like the Crazyflie AI-deck, enabling sophisticated onboard AI capabilities in miniature drones weighing under 50 grams \citep{mengozzi2024}.

Integrated Systems-on-Chip (SoCs) balance general-purpose computing flexibility with specialized AI acceleration capabilities (\citep{dicecio2024, valente2024, krger2025, garofalo2025}). The Qualcomm QRB5165, which powers platforms like the ModalAI VOXL2, integrates an 8-core CPU, Adreno GPU, Hexagon DSP with Tensor Accelerator, and PX4 flight controller support into a 16-gram package. This integration enables seamless execution of ROS2 nodes and OpenCV algorithms while maintaining real-time flight control. For more compute-intensive perception tasks, the Jetson Orin Nano (28g module) represents an optimal balance between computational capability and SWaP constraints, though even this advanced platform cannot yet achieve real-time semantic 3D Gaussian Splatting (\citep{lin2024, hong2025, li2025}) without substantial optimization efforts such as the GPU-specific strategies employed by GauRast \citep{li2025a}.

This hardware heterogeneity naturally leads to a distributed computational architecture where task placement aligns with hardware capabilities \citep{wang2023}. A representative implementation might employ a Zynq-based board as a real-time sensor hub managing low-level control loops, communicating via high-speed interfaces with a Jetson module executing perception and planning algorithms. This architecture aligns naturally with hierarchical reinforcement learning frameworks that decompose complex behaviors into layered abstractions.

For computational tasks exceeding edge platform capabilities \citep{xia2024, chandran2024}, fog computing architectures \citep{li2020, candalventureira2022} or cloud resources \citep{kim2024} connected via high-bandwidth wireless links---including 5G/6G networks \citep{geraci2022, cattai2025}, Wi-Fi 6E, or specialized drone communication systems like Droneforge Nimbus---enable advanced processing capabilities such as real-time 3D reconstruction. This distributed approach acknowledges a fundamental insight: not every computation must occur on the drone platform itself. Strategic workload offloading while maintaining critical functions onboard enables capabilities that would be impossible with pure edge computing, as demonstrated by edgeSLAM2's distributed SLAM architecture \citep{li2024}.

\subsection{Aerodynamics in Confined Spaces: The Invisible Challenge}

Indoor flight introduces complex aerodynamic phenomena rarely encountered in outdoor operations, fundamentally shaping platform control strategies (\citep{bartholomew2015, luo2024, lin2025}). The well-documented in-ground effect (IGE, \citep{he2020}) substantially augments thrust generation when operating near floors, but this benefit comes with significant pitch and roll instabilities requiring active compensation. Similar aerodynamic phenomena manifest near ceilings \citep{ai2024}, where low-pressure regions develop above the rotors, creating additional thrust augmentation but introducing control complexity. Operation near vertical surfaces induces wall effects \citep{paz2020} that generate powerful lateral forces and pitching moments exhibiting exponential relationships with proximity. These forces constitute dominant disturbances rather than minor perturbations, representing primary obstacles to achieving robust autonomy in confined spaces.

To maintain stability under these challenging conditions, modern indoor drones employ sophisticated control strategies that explicitly account for environmental interactions. Nonlinear Model Predictive Control (NMPC) approaches \citep{gupta2025, wang2025a} have demonstrated particular effectiveness by incorporating proximity effects directly into internal dynamic models, enabling prediction and compensation for aerodynamic disturbances at control frequencies exceeding 100 Hz. The development of computationally efficient solvers like TinyMPC \citep{nguyen2024} has made these sophisticated optimization algorithms feasible on resource-constrained embedded processors typical of small drones.

Complementing model-based approaches, Deep Reinforcement Learning (DRL) offers powerful capabilities for discovering complex nonlinear control policies that map high-dimensional sensor inputs directly to motor commands (\citep{kaufmann2023, mo2024, khan2025, romero2025}). Recent advances in autopilot software architecture, exemplified by the integration of TensorFlow Lite models directly into the PX4 flight stack \citep{hegre2025}, provide direct pathways for deploying learned policies from simulation environments to physical platforms, dramatically accelerating the development cycle for end-to-end neural control systems.

The most promising direction for achieving robust indoor autonomy lies in hierarchical hybrid architectures that synergistically combine multiple control paradigms \citep{reiterHierarchicalReinforcementLearning2024}. In these systems, high-level task-aware policies trained using DRL serve as strategic planners, determining semantic objectives such as ``inspect the ceiling corner for potential water damage indicators.'' These high-level goals are then translated by low-level NMPC-based controllers into dynamically feasible trajectories that respect hard safety constraints defined through Control Barrier Functions (CBFs, \citep{batool2025}). This layered approach combines the adaptive intelligence of learning-based methods with the mathematical guarantees provided by model-based control theory \citep{manzoor2024}.

Hardware innovations complement these software advances. Novel platforms like the Duawlfin drone \citep{tang2025} demonstrate how morphological computation---combining aerial and terrestrial locomotion modes---can physically circumvent challenging aerodynamic regimes by transitioning to ground movement when near-surface effects become problematic.

State estimation and uncertainty handling represent foundational challenges for any control system \citep{he2025}. Autonomous flight in GPS-denied indoor environments relies on fusing data from Inertial Measurement Units (IMUs), which suffer from integration drift, with visual sensors that can fail catastrophically in texture-less or poorly illuminated environments \citep{lin2024a}. This inherent uncertainty in position, orientation, and velocity estimates can rapidly destabilize control loops if not properly managed.

Consequently, a key research frontier involves developing adaptive state estimation systems that transcend fixed sensor models. Self-supervised frameworks that jointly optimize learned inertial odometry modules with differentiable MPC controllers \citep{he2025a} represent one promising direction. Equally important are sensor fusion strategies that dynamically adjust weighting based on real-time quality assessments \citep{irfan2025}, transforming state estimators from passive data integrators into active, environment-aware components essential for robust deployment.

The integration of large-scale AI models is driving the evolution of autonomous drones from reactive systems to proactive agents capable of environmental understanding and reasoning. Large Language Models (LLMs, \citep{chen2024a, cai2025}) and Vision-Language Models (VLMs, \citep{sun2025, blei2025, saxena2025}) provide semantic bridges enabling drones to interpret high-level mission commands and adapt behavior based on contextual understanding. For instance, identifying humans in the environment triggers increased safety margins and modified trajectory planning (\citep{cai2025, han2025, tadevosyan2025}).

Frameworks leveraging Retrieval-Augmented Generation (RAG) extend these capabilities by providing access to external knowledge bases containing building blueprints, maintenance histories, or material specifications \citep{batoolImpedanceGPTVLMdrivenImpedance2025a}. This culminates in Self-Adaptive Systems (SAS) capable of task-and-architecture co-adaptation (TACA), where drones dynamically reconfigure software architectures and control policies in response to environmental discoveries \citep{silva2025}.

\subsection{Collision-Tolerant Design: Embracing Contact}

The philosophy of collision tolerance represents a fundamental reconceptualization of drone design paradigms \citep{depetris2022, kumar2024}. Rather than treating surface contact as system failure, collision-tolerant architectures view gentle collisions as normal operational events during indoor navigation. This paradigm shift requires both mechanical innovations and computational adaptations \citep{depetris2024, yin2025}.

Recent advances have extended collision tolerance beyond passive protection to enable active contact-based sensing. Platforms like ScoutDI and Voliro demonstrate contact-based non-destructive testing (NDT) through specialized sensors including ultrasonic transducers, electromagnetic acoustic transducers (EMAT), and dry film thickness gauges \citep{watson2022}. These systems employ sophisticated control strategies---hybrid motion/force controllers and multi-state automata---ensuring stable surface contact despite environmental disturbances (\citep{santos2019, bodie2021, zhong2024, samadikhoshkho2025, piccina2025}).

Mechanically, shielded propeller designs with integrated bumpers provide practical collision protection. Carbon fiber or high-strength nylon shields combined with thermoplastic polyurethane (TPU) bumpers absorb impact energy while adding only 40-60 grams to platform weight. This minimal weight penalty enables drones to slide along walls during inspection, ensuring complete surface coverage. Computationally, Control Barrier Functions (CBFs) provide mathematical safety guarantees by defining admissible regions in state space that explicitly include gentle contact states (\citep{sharmaSafeControlDesign2023, tayalControlBarrierFunctions2024, chen2025}). Real-time optimization ensures control inputs maintain safety invariants while enabling aggressive maneuvers near surfaces.

Alternative architectural approaches serve different operational niches. Fully caged designs \citep{zhai2024} exemplified by the Flyability Elios series offer comprehensive collision protection and recovery capabilities through tumbling mechanisms, but incur weight penalties exceeding 200 grams while reducing sensor field of view. These platforms excel in extreme environments such as mines or industrial disaster sites \citep{deng2024, he2024}. Ducted fan coaxial designs \citep{siliang2024, shen2025} such as the Dronut X1 Pro provide inherent propeller protection with enhanced hover stability and reduced acoustic signatures, though they involve complex internal aerodynamics requiring sophisticated modeling. For residential and commercial indoor inspection applications, shielded quadcopters incorporating CBF-based control offer optimal balance between protection, design simplicity, maneuverability, and payload capacity.

\subsection{ROS2 as the Substrate for Modular, Semantic, and Scalable Indoor Autonomy}

\begin{figure}
    \vspace{-2pt} 
    \centering
    \includegraphics[width=1\linewidth]{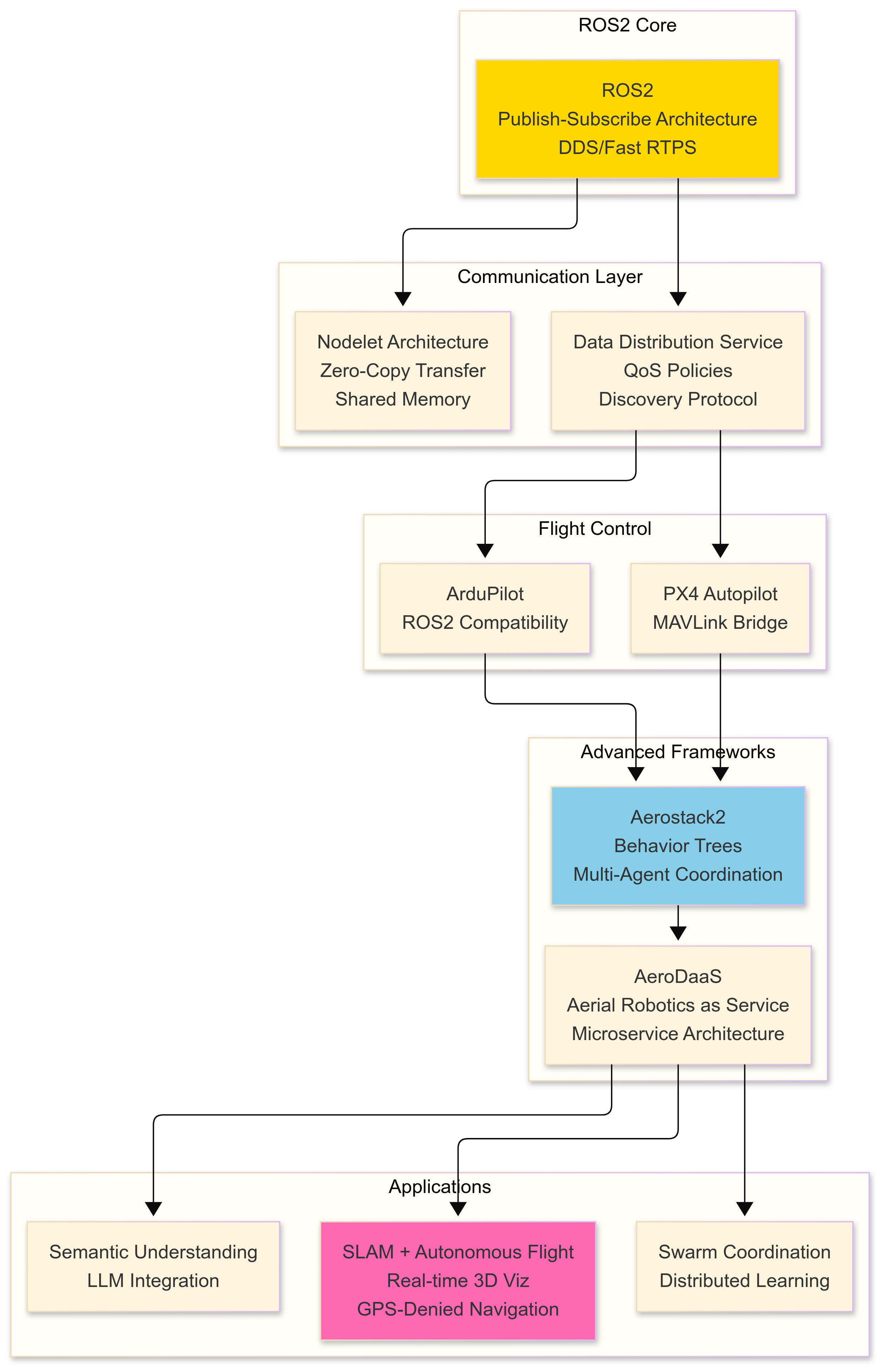}
    \caption{ROS2 provides architectural substrate for modular indoor autonomy through hierarchical organization from publish-subscribe middleware to semantic capabilities. DDS/Fast RTPS enables real-time communication; nodelets allow zero-copy multimodal processing. PX4/ArduPilot integration via MAVLink supports neural flight policies. Aerostack2 offers behavior trees and multi-agent coordination, culminating in AeroDaaS microservices. Enables: GPS-denied SLAM with sub-decimeter accuracy, distributed swarm learning, and LLM-based natural language mission specification.}
    \label{fig:ros}
\end{figure}

The transition to Robot Operating System 2 (ROS2) has fundamentally redefined the architectural baseline for embodied autonomy systems \citep{macenski2022}. Initially conceived as middleware for inter-process communication, ROS2 has evolved into a comprehensive abstraction layer enabling modular robotics development (\autoref{fig:ros}). Its publish-subscribe architecture, deterministic real-time capabilities, and extensive ecosystem of reusable components underpin advanced autonomy stacks across diverse aerial platforms \citep{maruyama2016}. In indoor inspection domains characterized by hardware heterogeneity, network constraints, and task diversity, ROS2's extensibility proves foundational \citep{thomas2014}.

Central to modern drone autonomy is seamless integration with flight control platforms. Both PX4 \citep{meier2015} and ArduPilot provide deep ROS2 compatibility, supporting not only traditional sensor fusion and control loops but increasingly accommodating neural network-based flight policies \citep{khawaldah2025, hegre2025}. These interfaces enable transparent bridging between high-level mission planning and low-level actuator control, facilitating tight integration of perception, planning, and action modules.

Advanced frameworks like Aerostack2 extend these capabilities by providing distributed behavior tree execution, multi-agent mission coordination, and swarm primitive libraries native to ROS2 \citep{fernandezcortizas2024}. These systems enable heterogeneous drone fleets to operate as coherent entities, with individual platforms executing modular mission primitives such as thermal anomaly detection, ceiling inspection, or 3D model upload. The abstraction of drone capabilities into microservices, formalized by the Aerial Robotics as a Service (AeroDaaS) model \citep{raj2025}, elevates ROS2 from communication middleware to a comprehensive service orchestration fabric.

This modular paradigm proves particularly powerful in indoor environments where GPS denial, spatial constraints, and mission-specific sensor payloads introduce both complexity and opportunity. ROS2's distributed communication infrastructure, built on Data Distribution Service (DDS) with Fast RTPS implementation \citep{pardocastellote2003}, supports sophisticated capabilities including federated learning pipelines with cryptographic security \citep{white2019}. This enables drones to locally train inspection models---for instance, defect detection from spectropolarimetric or thermal imagery---and synchronize learned parameters via bandwidth-efficient protocols without transmitting raw sensor data \citep{marino2023}. ROS2's nodelet architecture facilitates real-time execution of these learning algorithms onboard (\citep{gutirrez2018, casiniResponseTimeAnalysisROS2019, abaza2024, koudlil2025}), maintaining operational autonomy while preserving data security.

Semantic-level autonomy represents a significant advancement enabled by ROS2's architectural flexibility. Recent research in distributed swarm learning emphasizes sharing structured, high-level observations---such as ``suspected thermal anomaly detected behind drywall section''---rather than raw sensor streams \citep{an2025}. Integration with behavior tree frameworks \citep{colledanchise2018} and large language models enables natural language mission specification. Operators can issue commands like ``Scan all ceiling zones exhibiting depolarization ratios exceeding 0.8 and thermal deltas above 10\textdegree{}C,'' which individual agents translate into task-specific ROS2 computational graphs tailored to their sensor configurations and physical constraints.

Training these adaptive behaviors requires sophisticated simulation environments. Aerial Gym, built on NVIDIA's Isaac Gym, provides GPU-accelerated physics simulation enabling parallel training of inspection policies across thousands of simulated agents \citep{kulkarni2023}. While its ROS2 integration continues to mature, the platform enables training throughput orders of magnitude higher than traditional simulators, particularly for learning in cluttered indoor environments.

Gazebo, maintained by Open Robotics, offers native ROS2 integration with comprehensive sensor modeling capabilities \citep{koenig2004, ashwin2024}. The simulator provides high-fidelity models for RGB cameras, depth sensors (LiDAR, stereo, ToF), and IMUs with configurable parameters including field-of-view, resolution, update rates, and mounting positions \citep{jiang2025}. Critically, Gazebo incorporates realistic noise models---motion blur, thermal noise, IMU drift---essential for validating perception and control algorithms under realistic conditions. The simulated sensor outputs match what physical drones perceive during flight, enabling accurate testing of SLAM, obstacle avoidance, and semantic mapping pipelines before hardware deployment.

The organizational alignment between Open Robotics and Dronecode \cite{httpsdronecodeorg2025} ensures high fidelity between simulation and deployment. PX4, maintained under the Dronecode Foundation, facilitates interoperability across ROS2, MAVLink \citep{kouba2019, allouch2019}, and Gazebo through aligned standards and APIs. This organizational linkage creates a seamless simulation-to-deployment continuum where control logic and autonomy behaviors validated in Gazebo transfer predictably to PX4-powered hardware.

This architectural coherence manifests in commercial platforms like Hovermap \cite{jones2020}, which combines SLAM, autonomous flight, and real-time 3D visualization for GPS-denied indoor spaces. Hovermap exemplifies ROS2's orchestration potential: custom perception nodes for LiDAR-based loop closure and semantic labeling operate as ROS2 components communicating through DDS middleware; PX4 manages flight stability and mission execution; and visualization tools render the resulting data streams. These platforms demonstrate that ROS2 represents not merely a middleware choice but an architectural substrate coordinating sensing, learning, control, and operator interfaces into cohesive systems. The real-world success of platforms like Hovermap---operating in environments with no GPS, dynamic obstacles, and high semantic uncertainty---validates ROS2's role as the foundation for scalable, modular, and adaptive indoor autonomy.

While specialized tools like LidarView  offer high-performance point cloud rendering with SLAM playback capabilities, they lack the modular extensibility of ROS2-native tools like RViz \citep{groechel2022} or emerging alternatives like Rerun \cite{httpsrerunio2025} that integrate more naturally into autonomy development workflows. RViz is ROS's native 3D visualization tool deeply integrated with the ROS ecosystem for real-time robot monitoring and debugging \citep{abdulsaheb2023, kashyap2025}, while Rerun is a newer, developer-centric visualization platform designed for multimodal temporal data that offers faster performance through Rust implementation and supports programmatic data queries for robotics pipelines. Unlike RViz which requires ROS infrastructure and focuses primarily on visualizing a robot's sensor data, state, and environment, Rerun provides a more flexible SDK-based approach for logging and visualizing heterogeneous data streams across time, making it particularly suited for offline analysis, training workflows, and integration with modern AI/ML pipelines for Physical AI applications. Rerun has been used in projects such as HuggingFace Le Robot Project, Meta Reality Labs Aria, and a 3D Gaussian Splatting reconstruction engine Brush.

\subsection{The Sim-to-Real Challenge: From Virtual to Physical}

The convergence of physics-aware world models, embodied intelligence (\citep{long2025}; Yin et al.~2025; \citep{zhen2025}), and simulation-to-reality transfer represents a transformative advancement in UAV-based indoor inspection. While traditional datasets such as ScanNet++ \citep{yeshwanth2023} and Habitat-Matterport 3D \citep{yadav2023} emphasized visual appearance, the field now progresses toward physics-informed digital twins that capture dynamic interactions, causal relationships, and complex material behaviors essential for realistic training environments.

Recent simulation advances substantially accelerate this transformation. Ultra-fast physics engines like Genesis \cite{xian2024} achieve simulation speeds exceeding 43 million frames per second on single GPUs, while procedural world generators such as Infinigen Indoors \cite{raistrickInfinigenIndoorsPhotorealistic2024} enable automated creation of diverse, physically plausible environments with tunable characteristics. Frameworks like SOUS VIDE \citep{low2025} and UAVTwin \citep{choi2025} advance beyond traditional approaches by integrating 3D Gaussian Splatting (3DGS) directly into end-to-end visual policy learning pipelines. Unlike computationally intensive Neural Radiance Fields (NeRFs) \citep{miao2025, jeon2025}, 3DGS supports photorealistic simulation with real-time rendering capabilities---a critical requirement for interactive training scenarios.

The field of 3D indoor scene generation has undergone fundamental transformation through neural architecture adoption. DirectLayout \citep{ran2025} leverages large language models for spatial reasoning, generating coherent 3D layouts from textual descriptions. RoomCraft \citep{zhou2025} introduces multi-stage pipelines converting diverse inputs---images, sketches, or text---into complete 3D scenes through constraint-based optimization and conflict-aware object positioning. Hierarchical Scene Modeling (HSM) \citep{pun2025} addresses spatial scale challenges by generating realistic object arrangements from room-level furniture to desktop items. Meanwhile, LLplace \citep{yang2024} and OptiScene \citep{yang2025} demonstrate dynamic scene editing capabilities through fine-tuned open-source language models, advancing beyond static datasets toward customizable environments mirroring real-world complexity.

Integration with robotic simulation platforms like Gazebo enables comprehensive UAV policy training. VIPScene \citep{huang2024} employs video generation models to encode 3D physical knowledge, generating environments maintaining both visual coherence and structural consistency across multiple viewpoints---essential for drone navigation training. The capability to synthesize varied architectural styles and object densities addresses the sim-to-real gap by exposing UAVs to thousands of unique layouts before deployment. Scene Implicit Neural Fields \cite{liang2025} further enhance this by modeling multimodal relationships between spatial layouts and object arrangements, enabling generation of stylistically consistent environments grounded in real-world interior design principles.

Addressing data scarcity challenges, promising approaches generate 3D environments from ubiquitous 2D floor plans. As floor plan analysis capabilities mature---encompassing room boundary detection, classification, object recognition, and text extraction \citep{xu2025}---these documents provide rich semantic priors for scene synthesis. \citep{barreiro2023} employ differentiable rendering and adversarial training to reconstruct 3D models from single RGB floor plan images, while \citep{yin2024} extend this approach to construction monitoring by generating Building Information Models validated against point cloud data. These techniques prove particularly valuable for UAV-based reconstruction, enabling pre-training on diverse environments derived from architectural drawings.

In reinforcement learning contexts, differentiable simulation combined with 3DGS achieves remarkable zero-shot sim-to-real transfer capabilities, even for perceptually complex drone navigation tasks \citep{chen2025a}. These systems integrate multiple domain adaptation techniques: curriculum learning progressively increases task difficulty \citep{xingEnvironmentPolicyGenerative2025}, sophisticated sensor noise modeling captures real-world imperfections \citep{campanaro2023, mosleh2024}, and domain randomization extends beyond textures and lighting to material properties and physical parameters \citep{tiboni2023, ferede2025}. The result is robust visuomotor policies requiring no fine-tuning when deployed on physical platforms.

Modern simulation platforms such as NVIDIA Isaac Sim \cite{salimpour2025} begin supporting multimodal sensor modeling beyond traditional RGB-D, though critical modalities for property inspection---thermal, polarimetric, and hyperspectral imaging---remain unsupported \citep{nvidia2025, sim2025}. Physics-informed neural networks \citep{cuomo2022} offer pathways for modeling complex phenomena like moisture infiltration, material fatigue, and structural deformation---critical for restoration and appraisal applications where detecting latent damage proves essential.

Future inspection drones require simulation capabilities extending substantially beyond current offerings. Hyperspectral and polarimetric simulation pipelines would enable training models for material integrity assessment and moisture content determination---key requirements for UAD 3.6-compliant evaluations. UAVTwin \citep{choi2025} illustrates the importance of modeling realistic clutter and dynamic agents for insurance workflow training. Task-specific fusion algorithms---mapping degree of polarization across flooring to assess wear patterns or analyzing 1450nm short-wave infrared (SWIR) absorption bands to detect moisture in drywall---require drone-compatible simulation frameworks supporting these modalities.

End-to-end platforms like SOUS VIDE \citep{low2025} exemplify real-world robustness achievable through comprehensive simulation. Using only onboard computation and sensors, these systems demonstrate runtime adaptation to lighting variations, airflow disturbances, and dynamic occlusions---essential capabilities for disaster restoration contexts where environmental conditions prove unpredictable.

Ultimately, synthetic-to-real performance depends critically on training data quality and diversity. Beyond foundational datasets, future training pipelines must incorporate physically labeled scene variants with known spectral, polarimetric, and thermal ground truth for evaluating material substitutions and hidden damage detection. Multimodal digital twin libraries captured using specialized drone sensors should simulate temporal degradation patterns for longitudinal learning. UAD-annotated synthetic environments combining architectural semantics with condition-aware assets will support standardized pretraining for tasks like ANSI Z765-2021 floorplan generation \citep{miller2013} and Xactimate coding compliance \citep{burroughs2013}.

\subsection{Future Horizons: The Self-Improving Drone}

Recent advances in lifelong learning and agentic autonomy fundamentally reshape post-deployment adaptation capabilities for aerial robots (\citep{zhu2024, hernndez2025, zhang2025}). Rather than operating with static preprogrammed policies, modern drones increasingly incorporate mechanisms for continuous evolution through real-world experience, progressively refining internal models and behavioral strategies \citep{krger2025}.

Architectures such as Continual Reinforcement and Imitation Learning (CRIL) \citep{gao2021} and Continually Learning Prototypes \cite{hajizada2024} demonstrate how pseudo-rehearsal mechanisms and few-shot adaptation strategies effectively mitigate catastrophic forgetting in robotic systems. However, UAV-specific continual learning remains relatively underexplored compared to terrestrial manipulation tasks \citep{lesort2020}, presenting opportunities for domain-specific innovations. The scalability challenge for multi-agent systems has been addressed by frameworks such as that proposed by \cite{jiang2025a}, which introduce imitation learning methods enabling efficient path planning across fleets comprising tens of thousands of aerial robots.

Federated learning (FL) has emerged as a critical enabler for collaborative intelligence among autonomous systems, facilitating learning from decentralized data sources while preserving privacy and reducing communication overhead. Recent advances extend federated learning to cross-device scenarios where heterogeneous aerial robots collaboratively train shared models while maintaining data locality \citep{zhang2020, pham2021}. Sophisticated techniques including asynchronous model aggregation, trust-aware client selection, and lightweight model personalization \citep{qu2023, houUAVEnabledCovertFederated2023} address challenges posed by non-IID data distributions, intermittent connectivity, and constrained computational resources. Most significantly, newer approaches integrate continual learning with federated frameworks to support lifelong adaptation---enabling UAVs to incrementally update capabilities from local experiences while mitigating catastrophic forgetting through careful rehearsal strategies \citep{zhang2025a, tang2025a}.

Building upon these foundations of lifelong and federated learning, emerging forms of autonomy enable drones to interpret natural language mission specifications such as ``map this attic comprehensively for moisture infiltration risk indicators'' and autonomously execute end-to-end mission plans. These plans encompass sensor selection, active reconstruction strategies, and adaptive policy updates based on discoveries \citep{zhao2023, ravichandran2025}. Combined with sophisticated sim-to-real learning techniques \citep{salimpour2025, li2022}, this creates closed feedback loops where deployed agents inform next-generation simulation environments---transforming inspection workflows from static diagnostic procedures into dynamic, predictive systems.

As these methodologies mature, we envision indoor inspection drones that continuously adapt to specific building layouts through online learning, exchange knowledge via federated protocols \citep{zeng2020}, and autonomously identify failure cases for targeted retraining. While challenges remain in real-time model updating and standardizing cross-domain knowledge representations, the technical infrastructure for self-improving property intelligence systems rapidly approaches practical deployment readiness.

\section{Beyond Visual Perception: The Physics-Aware Sensing Stack}

The evolution of autonomous indoor drones from simple flying cameras to sophisticated inspection platforms depends fundamentally on their sensing capabilities. While early systems essentially adapted smartphone sensor suites to quadcopter platforms, the demands of professional property assessment require expanding perception beyond human visual limits. This transformation---from capturing surface appearance to understanding material composition and physical condition---represents a critical frontier at the intersection of robotics, optics, and computational imaging.

\subsection{The Foundation: Traditional ``Flying iPhone''}

The baseline sensor configuration for modern indoor drones consists of three core components: RGB camera, depth sensor, and inertial measurement unit (IMU). This trinity enables visual-inertial SLAM (Simultaneous Localization and Mapping) for basic navigation and 3D reconstruction, forming the foundation upon which more sophisticated capabilities are built \citep{xu2024, yarovoi2024}.


\subsubsection{RGB Cameras}

RGB Cameras serve as the primary visual perception system, providing rich texture information essential for feature tracking and photorealistic reconstruction. For property assessment applications conforming to standards like UAD 3.6, which mandates measurements accurate to the nearest inch with computer-generated floor plans showing detailed calculations, camera specifications become critical. Global shutter sensors  eliminate the rolling shutter artifacts that plague consumer cameras during rapid platform movements (\citep{seiskari2025, wu2025, mcgriff2025, carmichael2025}) (see \autoref{tab:shutter}), while wide-angle lenses must carefully balance field-of-view requirements against geometric distortion that complicates accurate 3D reconstruction (\citep{wu2025, deng2025, tiradogarin2025}).

\begin{table*}[htbp]
\centering
\caption{Comparison of global shutter and rolling shutter cameras for drone-based indoor inspection applications. Global shutter sensors eliminate motion artifacts critical for rapid UAV movements but require higher cost and power consumption, while rolling shutter options offer higher resolutions and lower power draw at the expense of image quality during dynamic flight.}
\label{tab:shutter}
\footnotesize
\begin{tabular}{@{}p{0.18\textwidth} p{0.20\textwidth} p{0.20\textwidth} p{0.35\textwidth}@{}}
\toprule
Parameter & Global Shutter & Rolling Shutter & Key Considerations \\
\midrule
Shutter Type & All pixels exposed simultaneously & Line-by-line exposure & Global shutter sensors eliminate the rolling shutter artifacts that plague consumer cameras during rapid platform movements \\
Motion Artifacts & None & Skew, wobble, partial exposure & Critical for drone applications with rapid movements \\
Cost & Higher (\$200-1000+) & Lower (\$50-200) & Global shutter typically 2-5x more expensive \\
Resolution Options & 1-12MP typical & 2-48MP common & Higher resolutions available in rolling shutter \\
Frame Rate & 30-120fps @ 1080p & 30-60fps @ 1080p & Global shutter often has higher frame rates \\
Power Consumption & Higher (200-500mW) & Lower (100-300mW) & Global shutter requires more complex circuitry \\
Communication & CSI-2, USB3, GigE & CSI-2, USB2/3 & CSI-2 provides lowest latency for embedded systems \\
Field of View & Wide-angle lenses must carefully balance field-of-view requirements against geometric distortion that complicates accurate 3D reconstruction & Same consideration & 60-120\textdegree{} typical, distortion correction critical \\
\bottomrule
\end{tabular}
\end{table*}

\subsubsection{Depth Sensing}

Depth Sensors provide the geometric scaffolding essential for creating accurate digital twins of indoor spaces. The selection between competing technologies---LiDAR versus Time-of-Flight (ToF)---represents a critical design decision with cascading implications (\autoref{tab:depth-sensors}). LiDAR systems exemplified by the Livox Mid-360 (\citep{ren2025, wu2025a, xu2025a, felix2025}) offer superior range accuracy and precision but impose significant weight and power penalties. Modern direct ToF (dToF) sensors utilizing single-photon avalanche diode (SPAD) technology, such as the ST VL53L5CX \citep{niculescu2024} and Sony AS-DT1, are progressively closing this performance gap while maintaining specifications compatible with small drone platforms. While indirect ToF (iToF) sensors using continuous-wave modulation offer lower cost and simpler implementation, they suffer from fundamental limitations including multipath interference, limited range accuracy, and higher power consumption compared to dToF systems, making them less suitable for precision indoor mapping applications \citep{gutierrez-barraganIToF2dToFRobustFlexible2021}.

\begin{table*}[htbp]
\centering
\caption{Comparison of depth sensing technologies for autonomous indoor inspection drones. Each technology presents distinct tradeoffs between accuracy, weight, power consumption, and operational characteristics that fundamentally shape sensor selection for UAV platforms. The weight-performance relationship emerges as the critical constraint, with direct Time-of-Flight (dToF) arrays offering the most promising balance for indoor inspection applications despite reduced range compared to LiDAR alternatives.}
\label{tab:depth-sensors}
\footnotesize
\begin{tabular}{@{}p{0.12\textwidth} p{0.17\textwidth} p{0.20\textwidth} p{0.20\textwidth} p{0.20\textwidth}@{}}
\toprule
Technology & LiDAR (e.g., Livox Mid-360) & Stereo RGB & dToF Arrays (e.g., Sony AS-DT1) & Plenoptic Cameras \\
\midrule
Accuracy & Centimeter-level accuracy & 1-5\% of distance & \textpm{}5cm accuracy at 10m & Sub-millimeter accuracy at close range (\textless2m) \\
Range & 70m & 0.5-20m & 40m indoor & Best \textless2m, 3\% error at 30-100m \\
Weight & 265g & Minimal (uses existing cameras) & 50g (announced specification) & 200-500g \\
Power & 14W peak & Low (processing only) & 1-3W & 5-10W \\
FoV & 360\textdegree{} horizontal, 59\textdegree{} vertical & Depends on baseline & \textgreaterequal{}30\textdegree{} horizontal & Limited by optics \\
Resolution & 100k-300k points/sec & Matches RGB resolution & 8\texttimes{}8 to 64\texttimes{}64 typical & Depends on microlens array \\
Environmental Robustness & Excellent & Poor in low texture & Good & Moderate \\
Computational Load & Low & High (stereo matching) & Low & Substantial computational requirements \\
\bottomrule
\end{tabular}
\end{table*}

Zero-shot metric depth estimation models like Metric3Dv2 \citep{hu2024} have gained popularity for predicting real-world distances from single images without domain-specific training \citep{xu2025b}, achieving state-of-the-art performance through canonical camera space transformation and training on 16+ million images. These models excel in non-metrological applications like 3D content creation and AR/VR, where relative depth and generalization matter more than absolute precision \citep{lin2025b}. However, they face fundamental metrological limitations from monocular vision's inherent scale ambiguity---without camera intrinsics or external references, these models cannot provide absolute metric accuracy as identical 2D projections map to multiple valid 3D reconstructions (\citep{marsalFoundationModelsMeet2024, birkl2023, cai2024}).

Zero-shot metric depth estimation models have proven valuable for enhancing stereo RGB depth estimation, particularly in challenging scenarios with occlusions and textureless regions \citep{cheng2025}. These models have also shown promise in augmenting sparse ToF measurements by leveraging ToF sensors for metric constraints while providing dense, high-resolution depth maps. Notable implementations include Prompt Depth Anything, which uses LiDAR as a metric prompt \citep{lin2025b}, and systems like DEPTHOR \citep{xiang2025} and SelfToF \citep{ding2025} that integrate consumer-grade 8\texttimes{}8 VL53L5CX dToF arrays. Beyond these hybrid approaches, the field is witnessing broader computational advances in sparse ToF sensing, including multi-frame integration techniques \citep{conti2025} and time-resolved sensing capabilities enabled by SPAD arrays \citep{teja2025}. Together, these developments promise to blur traditional boundaries between sensing modalities, demonstrating how computational techniques can dramatically enhance the capabilities of low-cost, lightweight sensors.

The weight-performance tradeoff fundamentally shapes sensor selection for UAV applications. The Livox Mid-360 provides comprehensive 360\textdegree{} horizontal and 59\textdegree{} vertical field coverage with centimeter-level accuracy, but its 265g weight and 14W peak power consumption represent substantial burdens for platforms where every gram directly impacts flight duration. In contrast, the Sony AS-DT1 achieves a paradigm shift at just 50g (announced specification; some sources indicate 34g in development variants), delivering \textpm{}5cm accuracy at 10m through dToF SPAD technology. This five-fold weight reduction comes with expected tradeoffs: reduced maximum range (40m indoor versus 70m) and narrower field of view (\textgreaterequal{}30\textdegree{} horizontal).

Plenoptic cameras represent an alternative approach, capturing complete light field information through microlens arrays to enable post-capture refocusing and extended depth of field. Industrial models like those from Raytrix achieve sub-millimeter accuracy at close range (\textless2m), but their 200-500g weight, substantial computational requirements, and degraded accuracy at typical UAV operating distances (3\% error at 30-100m) present significant implementation challenges. Despite these limitations, their unique capability to capture both 3D depth and 2D imagery in a single exposure without moving parts makes them valuable for specialized applications where refocusing capabilities and dense depth maps outweigh weight penalties.

The hovering instability inherent to UAV platforms introduces additional complexity, causing depth sensing errors proportional to localization uncertainty. Research on UAV depth error mitigation \citep{wang2023a, wang2023a} demonstrates that active sensors like ToF and LiDAR maintain superior accuracy during platform motion compared to passive optical methods. Automotive LiDAR systems have addressed similar vibration challenges through MEMS mirror stabilization and adaptive signal processing, with studies showing timing jitter from vibration can be reduced through FPGA-based real-time compensation. In satellite remote sensing, micro-vibration compensation employs both passive isolation through Stewart platforms and active Line-of-Sight stabilization with Inertial Reference Units, achieving sub-pixel stability for multispectral imaging at orbital velocities \citep{tang2020}.

The challenge intensifies dramatically at close indoor ranges where sub-millimeter platform motion represents a significant fraction of working distance. Studies on optical flow sensors demonstrate accuracy degradation from 97\% to 92\% when transitioning from stationary to UAV-mounted configurations at 0.3-2m distances \citep{baiStructuralVibrationDetection2025}. Unlike automotive applications benefiting from vehicle mass stabilization or satellite platforms with predictable orbital mechanics, small UAVs face stochastic disturbances from rotor downwash, air currents, and control loop oscillations.

\subsubsection{IMUs}

Inertial Measurement Units (IMUs) provide high-frequency motion estimates critical for maintaining stable flight and enabling sensor fusion. While consumer-grade MEMS IMUs like the TDK ICM-42688 suffice for basic stabilization requirements \citep{usv2025}, professional inspection applications demand careful attention to drift characteristics, calibration stability, and noise performance (\autoref{tab:imu-comparison}). Recent research directions include IMU array configurations for improved accuracy \citep{will2024}, sophisticated uncertainty modeling techniques \citep{qiu2024}, data-driven learning approaches for drift compensation (\citep{qiu2025, altawaitan2025, he2025a}), and generative models for IMU data augmentation \citep{zhou2025a}.

Tightly-coupled LiDAR-Inertial-Camera fusion represents the current state-of-the-art for robust SLAM in inspection UAVs, addressing single-sensor limitations through complementary modality integration. Recent implementations leverage 3D Gaussian Splatting for photorealistic reconstruction while maintaining real-time performance. \citep{wang2024a} achieve this by initializing Gaussian primitives from both triangulated visual features and colorized LiDAR points, effectively compensating for sensor blind spots common in indoor environments. The IMU's contribution extends beyond simple dead reckoning: continuous-time trajectory optimization frameworks \citep{li2025c} exploit high-frequency inertial data to maintain accurate pose estimates during aggressive maneuvers where visual tracking fails, while factor graph formulations \citep{zhang2025b} enable pyramid-based training of multi-level features robust to motion blur.

For inspection robots experiencing intense vibrations, such as bionic quadrupeds, \citep{chen2025b} demonstrate that adaptive graph inference in visual-inertial odometry maintains accuracy where traditional loosely-coupled approaches fail catastrophically. The integration of zero-shot depth models \citep{li2025c} further enhances geometric accuracy by combining sparse LiDAR measurements with RGB appearance cues, proving particularly valuable for inspection scenarios involving reflective surfaces or limited sensor coverage.

While this traditional sensor configuration has proven effective for basic mapping tasks, it faces fundamental limitations in property assessment contexts. Visual SLAM  (Simultaneous Localization and Mapping) algorithms fail in texture-less environments common in unfinished buildings or minimalist interiors \citep{dong2023, maQualiSLAMQualityDrivenFeature2025}. Depth sensors struggle with reflective surfaces ubiquitous in modern construction---windows, mirrors, and polished fixtures (\citep{tan2023, huang2023, shen2025a}). Most critically, these sensors capture only geometric structure and visual appearance---they cannot distinguish genuine marble from convincing polymer imitations, detect moisture accumulation behind painted walls, or quantify material degradation invisible to human perception.

\begin{table*}[htbp]
\centering
\caption{Comparison of Inertial Measurement Unit (IMU) technologies for autonomous indoor drones. Consumer MEMS devices provide adequate performance for basic stabilization, while professional-grade units offer order-of-magnitude improvements in noise characteristics and bias stability critical for precise indoor navigation. IMU arrays represent an emerging cost-effective approach that achieves intermediate performance through statistical averaging of multiple low-cost sensors.}
\label{tab:imu-comparison}
\footnotesize
\begin{tabular}{@{}p{0.15\textwidth} p{0.20\textwidth} p{0.16\textwidth} p{0.18\textwidth} p{0.20\textwidth}@{}}
\toprule
Type & Consumer MEMS (e.g., ICM-42688) & IMU Arrays & Professional Grade (e.g., VN-100, ADIS16495) & Key Parameters \\
\midrule
Gyroscope Noise & 0.005-0.01\textdegree/s/$\sqrt{\text{Hz}}$ & Reduced by $\sqrt{N}$ & 0.001-0.005\textdegree/s/$\sqrt{\text{Hz}}$ & Lower is better \\
Accelerometer Noise & 50-100 $\mu$g/$\sqrt{\text{Hz}}$ & Reduced by $\sqrt{N}$ & 10-50 $\mu$g/$\sqrt{\text{Hz}}$ & Critical for pose estimation \\
Bias Stability & 1-10\textdegree/hr & Improved through averaging & 0.1-1\textdegree/hr & Affects long-term drift \\
Update Rate & 100-8000 Hz & Same as individual & 100-1000 Hz & Higher enables better fusion \\
Power & 5-10 mW & $N \times$ single IMU & 20-50 mW & Arrays scale linearly \\
Cost & \$5-20 & \$50-200 & \$600-4000 & Arrays offer cost-effective improvement \\
\bottomrule
\end{tabular}
\end{table*}

\subsection{Physical Sensing: Beyond Appearance to Material Truth}

The evolution from conventional visual sensing to advanced physical sensing modalities represents a fundamental shift in how autonomous systems perceive and understand their environment. While traditional RGB cameras capture what the human eye sees---surface appearance and color---advanced sensing technologies reveal the underlying physical and chemical properties that determine a structure's true condition. By analyzing how materials interact with different wavelengths and polarizations of light, these modalities expose information invisible to conventional imaging, enabling quantitative assessment of material composition, surface physics, and structural integrity.

\subsubsection{Hyperspectral Imaging: Chemical Fingerprinting}

Hyperspectral imaging (HSI) transcends the limitations of RGB cameras by revealing what materials fundamentally are, rather than merely how they appear. While conventional cameras capture three broad spectral bands mimicking human vision, HSI systems record hundreds of narrow, contiguous spectral bands across the electromagnetic spectrum (\citep{salcido2024, valme2025, mukhtar2025}). This spectral resolution enables material identification through their unique ``fingerprints''---characteristic patterns of absorption and reflection determined by molecular structure. Where a conventional camera sees simply ``brown wood floor,'' HSI can definitively distinguish genuine oak hardwood from polymer-based laminate designed to mimic it, a distinction with significant implications for property valuation and authenticity verification.

Recent advances in deep learning have dramatically enhanced HSI's classification capabilities. \citep{sifnaios2024} demonstrated near-perfect accuracy in material classification using neural networks trained on hyperspectral data, while \citep{zahiri2021} established HSI's superiority over multispectral systems for building material characterization---a critical capability for comprehensive property assessment.

For indoor inspection applications, HSI offers several transformative capabilities. Water detection represents perhaps the most valuable application, as water exhibits strong absorption features in the short-wave infrared (SWIR) spectrum, particularly at 1450nm and 1930nm \citep{levy2014}. This enables HSI-equipped systems to create precise moisture maps revealing hidden water damage before it manifests as visible stains or structural degradation \citep{rath2023, li2025d}. Beyond moisture detection, HSI enables early-stage paint degradation assessment by identifying chemical breakdown in pigments and binders \citep{ma2023a}, fungal growth detection through species-specific spectral signatures before visible symptoms appear \citep{kristensen2023}, and hazardous material identification including asbestos fibers \citep{bonifazi2019}.

While commercial systems like the Specim AFX platform have demonstrated HSI's maturity in outdoor applications, translating this capability to indoor drone platforms faces significant miniaturization challenges. Traditional hyperspectral instruments rely on complex optical assemblies---dispersive elements, multiple detectors, and precise mechanical components---that result in systems weighing several kilograms. The challenge of packaging this functionality into sub-500g platforms suitable for indoor drones represents a classic adoption barrier: market demand awaits technical feasibility while development investment requires proven demand.

The breakthrough enabling drone-compatible HSI lies in reimagining optical systems through computational imaging and metasurface technology. Metasurfaces---ultrathin arrays of subwavelength nanostructures---manipulate light in ways impossible with conventional optics (\citep{lin2023, hu2024a, zhang2025c}). Rather than using bulky prisms or gratings to disperse light, metasurfaces encode spectral information through carefully designed nanostructure patterns. \citep{lin2023} demonstrated snapshot hyperspectral imaging using a single multi-wavelength metasurface chip requiring minimal training data---a crucial advantage for resource-constrained platforms. Similarly, \citep{zhangRealtimeMachineLearning2024} achieved real-time spectral imaging by combining encoding metasurfaces with neural networks, while \citep{hu2024a} outlined comprehensive frameworks for metasurface-based computational imaging suitable for ultra-compact form factors. These innovations fundamentally shift complexity from hardware to software, potentially shrinking briefcase-sized spectrometers to components no larger than smartphone camera modules.

Yet even with revolutionary miniaturization, practical deployment in indoor environments presents unique challenges distinct from outdoor applications. Indoor spaces feature complex, heterogeneous lighting conditions---mixtures of fluorescent tubes, LED panels, incandescent bulbs, and natural light from windows---that complicate spectral analysis \citep{davies2020, yuan2025}. Each light source has its own spectral distribution, creating a patchwork of illumination that varies spatially and temporally. \citep{moghadam2021} developed data-driven methods to enable indoor HSI under these challenging conditions, using machine learning to separate material properties from illumination effects. \citep{morimoto2020} characterized typical indoor illumination patterns to inform correction strategies, creating databases of common lighting scenarios.

Another significant challenge arises from the angular dependence of spectral reflectance---a material's spectral signature changes based on the angles of observation and illumination. This necessitates either sophisticated geometric corrections using depth sensor data to account for surface orientation, or development of view-invariant spectral descriptors that remain consistent across different viewing angles \citep{kizel2023, yuan2024}. Some researchers are exploring adaptive illumination strategies where drones dynamically adjust their onboard lighting based on the material being inspected and the geometric configuration \citep{tang2023}. These active illumination approaches, while adding system complexity, provide the consistent, broad-spectrum light necessary for reliable spectral measurements in variable indoor conditions.

Beyond physical miniaturization, spectral calibration presents another fundamental challenge for indoor HSI deployment \citep{andersen2026}.. The initial calibration uncertainties from manufacturing tolerances compound through environmental drift, where temperature fluctuations induce material thermal expansion causing wavelength shifts up to 0.0016 nm per degree Celsius---potentially leading to several nanometer errors under typical indoor temperature variations. When factory calibration alone proves insufficient, dynamic recalibration (online) strategies need to be implemented. Inter-device calibration transfer methods enable model sharing between different HSI systems without full recalibration \citep{yang2022}. Deep learning approaches show particular promise for use cases with not so stringent requirements for metrologically correct spectrum, with calibration-invariant neural networks \citep{botton2019}.

The criticality of accurate spectral calibration becomes evident when considering that spectral reflectance serves merely as a proxy for the ultimate goal: spectral unmixing for material characterization \citep{haijen2025}. Modern unmixing algorithms like UnMix-NeRF integrate spectral unmixing into neural radiance fields, learning global endmember dictionaries that represent pure material signatures---but calibration errors propagate nonlinearly through these algorithms, corrupting both diffuse and specular component estimation \citep{perez2025}. Even small calibration errors of 1-2 nm can cascade through spectral matching algorithms, causing material misidentification \citep{andersen2026}. Physically accurate spectral reflectance measurement remains essential for leveraging existing spectral libraries, as cross-platform compatibility and temporal consistency require absolute rather than relative calibration \citep{rastiImageProcessingMachine2024}. Thus, while miniaturization enables drone deployment, the spectral calibration challenge demands sophisticated computational approaches that transform fixed hardware limitations into software-solvable problems, ensuring that HSI can deliver on its promise of revealing the invisible chemistry of indoor environments.

Despite these implementation challenges, HSI represents one of the few sensing modalities capable of revealing the invisible---offering early detection of moisture intrusion, material degradation, and biological contamination that would otherwise remain hidden until costly damage occurs. When integrated with polarimetric imaging (which reveals surface physics rather than chemistry) and traditional RGB-D sensing for geometric context, HSI forms part of a powerful multimodal sensing suite. Together with established non-destructive testing (NDT) techniques \citep{masri2020, zhang2025d}, these technologies can transform drones from flying cameras into comprehensive material analysis laboratories. The path to commercial viability requires continued innovation in computational imaging algorithms that compensate for hardware constraints, development of standardized material spectral libraries for reliable field deployment, and integration frameworks that seamlessly combine hyperspectral data with other sensing modalities. As these technical and practical challenges are addressed, HSI-equipped drones promise to revolutionize property assessment by shifting from subjective visual inspection to objective, physics-based evaluation.

\subsubsection{Polarimetric Imaging: Surface Physics Revealed}

While hyperspectral imaging reveals material composition through chemical signatures, polarimetric imaging exposes physical surface properties and conditions by analyzing how materials modify the polarization state of reflected light. This complementary modality provides information invisible to both conventional cameras and hyperspectral sensors, quantifying surface roughness, detecting mechanical stress, and enhancing depth perception in challenging scenarios \citep{dubovik2019, stock2020}.

The mathematical foundation of polarimetric imaging rests on the Stokes-Mueller formalism, which provides a complete description of polarized light and its interaction with materials. Four Stokes parameters (\(S_0\), \(S_1\), \(S_2\), \(S_3\)) describe the polarization state of light, encoding total intensity, linear polarization components, and circular polarization. When light interacts with a material, the 4\texttimes{}4 Mueller matrix \(M\) characterizes how that material transforms the polarization state through the relationship $S_{\text{out}} = M \cdot S_{\text{in}}$. This physics-based approach provides quantitative, verifiable data essential for standards compliance and objective assessment.

Polarimetric imaging excels at quantifying surface conditions that affect material integrity and appearance. Surface roughness assessment represents a primary application \citep{han2023}---smooth surfaces tend to maintain the polarization state of incident light, while rough surfaces depolarize it through multiple scattering events. This enables objective condition assessment through metrics like the Linear Polarization Ratio (LPR) and Degree of Linear Polarization (DoLP) \citep{hall2013}. For comprehensive material characterization, the full Mueller matrix captures complex phenomena including depolarization from volumetric scattering within materials like concrete and composites---situations where simpler Jones calculus fails due to its inability to handle partially polarized light.

Beyond material characterization, polarimetric imaging significantly enhances depth estimation and 3D reconstruction capabilities, particularly in scenarios where traditional stereo vision struggles. Shape from Polarization (SfP) techniques \citep{yoshidaImprovingTimeofFlightSensor2018} exploit the relationship between surface orientation and polarization patterns to recover surface normals. This enables depth reconstruction even for textureless surfaces, transparent materials, or highly reflective surfaces where conventional stereo matching fails due to lack of features or specular reflections \citep{huang2023}. Recent advances demonstrate the power of fusing polarimetric data with traditional depth sensors \citep{ikemura2024}, using polarization-derived surface normals to refine and enhance depth estimates in challenging environments.

For robotic inspection in built environments \citep{taglione2024}, polarimetric imaging addresses several critical challenges. The physics of polarized light enables robust discrimination between materials that appear visually similar but have different surface properties. Conductors and dielectrics, for instance, show distinctly different polarimetric signatures due to differences in their complex refractive indices \citep{li2024a}. This capability proves invaluable for material authentication---genuine hardwood's aligned cellulose fibers produce optical anisotropy that creates characteristic polarization patterns, while synthetic laminates typically yield isotropic responses beneath their melamine coatings. Mueller matrix decomposition can quantify subsurface defects through depolarization indices and detect mechanical stress through retardance parameters, enabling predictive maintenance before visible failures occur.

The controlled nature of indoor environments paradoxically complicates passive polarimetric approaches. Unlike outdoor scenes where sunlight provides naturally polarized illumination at varying angles throughout the day, indoor artificial lighting typically lacks the angular diversity needed for comprehensive polarimetric analysis \citep{zhan2019}. This limitation has driven development of active illumination strategies where drones carry their own polarized light sources. Illuminating surfaces at the Brewster angle---where reflected light becomes maximally polarized---dramatically enhances material contrast. This can be implemented through gimbaled LED arrays that adjust their angle based on surface orientation. Advanced systems employ reinforcement learning algorithms to optimize illumination patterns in real-time, learning that specific material pairs like genuine marble versus faux marble finishes exhibit maximum polarimetric contrast at particular geometric configurations.

Recent computational advances have elegantly addressed hardware limitations that previously restricted polarimetric imaging to laboratory settings. Traditional Mueller matrix polarimetry requires 16 sequential measurements with different polarizer configurations---impractical for moving platforms. Modern snapshot division-of-focal-plane (DoFP) sensors capture multiple polarization states simultaneously but typically only measure linear Stokes parameters. Deep neural networks now bridge this gap, reconstructing full Mueller matrices from partial DoFP data \citep{chae2024}, effectively upgrading commercial sensors through sophisticated software \citep{yang2024a}. This computational approach shifts the burden from exotic hardware to sophisticated processing pipelines that include real-time denoising of single-shot polarimetric data \citep{hu2020, li2020a}, physics-informed neural networks for matrix completion that respect fundamental constraints, and end-to-end models that directly map polarimetric signatures to industry-standard condition metrics.

The strategic pathway forward for polarimetric sensing emphasizes data-centric approaches: building comprehensive databases of polarimetric material properties and training robust AI models for real-world deployment \citep{pereira2024, lincetto2025}. Proprietary databases of polarimetric bidirectional reflectance distribution functions (pBRDFs) for common building materials create significant competitive advantages. When combined with RGB-D sensing for geometric context \citep{kadambi2015} and hyperspectral data for chemical identification, polarimetric imaging enables creation of queryable physical digital twins that encode not just geometry but quantitative material states---fundamentally transforming how autonomous systems perceive and understand built environments.

\subsubsection{The Power of Fusion: Spectropolarimetric Sensing}

While hyperspectral and polarimetric imaging each provide powerful insights into material properties, their true transformative potential emerges when combined into spectropolarimetric imaging---simultaneously capturing both spectral and polarimetric information across multiple wavelengths. This fusion creates a sensing modality that transcends the limitations of its individual components, revealing material properties and conditions impossible to detect with either technique alone. Where hyperspectral imaging might identify the presence of moisture through water's characteristic absorption bands, spectropolarimetry can additionally determine whether that moisture has penetrated a protective surface coating or remains on top, based on wavelength-dependent depolarization signatures \citep{jiang2024}. For property inspection, this distinction between harmless surface condensation and destructive water infiltration could prevent billions in misdiagnosed damage claims annually.

The mathematical elegance of spectropolarimetric imaging lies in capturing the full wavelength dependence of the Mueller matrix M($\lambda$), where each of the 16 matrix elements varies with wavelength according to the material's dispersive properties \citep{wangPhysicalInterpretationMueller2017}. This spectral Mueller matrix contains orders of magnitude more information than single-wavelength polarimetry or spectroscopy alone. Consider stress detection in glass: while standard polarimetry might detect stress presence through optical retardance, spectropolarimetric analysis reveals the complete stress-optical dispersion curve. This enables not only quantitative stress magnitude determination but also discrimination between thermal and mechanical stress origins based on their distinct spectral signatures \citep{oka2003}. Similarly, the wavelength-dependent diattenuation of wood varies systematically with lignin and cellulose content ratios, potentially enabling assessment not just of wood authenticity but also quality, age, and treatment condition.

For indoor robotic inspection aligned with industry standards, spectropolarimetric imaging enables unprecedented diagnostic capabilities that could revolutionize property assessment. Consider comprehensive water damage evaluation: water exhibits characteristic absorption features at 1450nm and 1930nm in the near-infrared spectrum. When water penetrates porous materials like gypsum drywall, it creates complex wavelength-dependent scattering phenomena that manifest as increased depolarization specifically at these water absorption wavelengths. A spectropolarimetric system could map not just moisture presence but also penetration depth, flow patterns, and the material degradation state---transforming subjective damage categories into quantitative, defensible metrics. For material authenticity verification crucial to accurate property valuation, the combination of spectral signatures (revealing chemical composition) with polarimetric signatures (revealing surface microstructure and bulk properties) creates a dual authentication system virtually impossible to counterfeit with surface treatments alone.

However, implementing spectropolarimetric imaging on weight-constrained drone platforms faces formidable technical challenges. Traditional approaches would require either sequential acquisition through motorized filter wheels---incompatible with platform motion and vibration---or parallel beam-splitting architectures with multiple sensors that quickly exceed weight budgets. The data volume alone presents a staggering challenge: a full spectropolarimetric datacube with 100 spectral bands and 16 Mueller matrix elements per pixel generates approximately 533 times more data than a standard RGB image, overwhelming both onboard storage and wireless transmission capabilities of current drone systems.

The breakthrough enabling drone-compatible spectropolarimetric imaging emerges from innovative snapshot computational imaging solutions (\citep{mu2022, han2023a, dai2023, wen2025}). Recent demonstrations show that metasurfaces can encode both spectral and polarimetric information onto a single focal plane through carefully designed nanostructure arrays that manipulate light in multiple dimensions simultaneously. The key innovation lies in designing meta-atoms with coupled spectral and polarimetric responses: chiral nanostructures that exhibit both wavelength-dependent transmission and circular dichroism, effectively encoding multiple types of information in a single optical interaction \citep{basiri2019}.

One promising architecture for practical implementation employs a ``multimodal super-pixel'' design where each logical pixel contains an array of diversely functionalized meta-atoms \citep{quMultimodalLightsensingPixel2022}. Within a 4\texttimes{}4 meta-atom group, different elements might be optimized for specific wavelength/polarization combinations, creating an overcomplete basis set for computational reconstruction. Machine learning algorithms, particularly physics-informed neural networks that incorporate Maxwell's equations as constraints, then decode the complex intensity pattern captured by the sensor to recover the full spectropolarimetric datacube \citep{zhangRealtimeMachineLearning2024}. This computational approach fundamentally shifts complexity from hardware to software, enabling sub-100g implementations suitable for integration into inspection drones.

Real-world applications demonstrate the transformative potential of spectropolarimetric sensing for building inspection. The presence of moisture induces fundamental alterations in both the physical and chemical properties of building materials---modifications readily detectable through spectropolarimetric analysis. In porous materials such as wood and concrete, moisture content demonstrates direct correlation with both spectral reflectance and polarization signatures \citep{dereniak2003}. Wood substrates exhibiting moisture content exceeding the critical 18-20\% threshold become susceptible to biological decay processes, with associated compositional changes manifesting as distinctive alterations in near-infrared spectral signatures. Analogously, concrete structures are considered compromised when moisture content surpasses 3.5-4.5\%, a condition detectable through combined spectral-polarimetric analysis \citep{tramex2025}. Recent advances in spectropolarimetric datasets \citep{jeon2024} and neural spectro-polarimetric field representations \citep{kim2023} enable sophisticated correlation between measured data and established moisture damage signatures, facilitating a paradigm shift from binary qualitative assessments to precise quantitative determination.

The commercial implications of widespread spectropolarimetric adoption are substantial. Contemporary property assessment methodologies remain largely dependent on subjective visual inspection protocols, frequently resulting in valuation disputes and inconsistent damage assessments. Implementation of drone-mounted spectropolarimetric systems promises to generate comprehensive documentation with unprecedented precision and objectivity. Instead of subjective observations like ``water staining observed on ceiling,'' inspectors could provide quantitative assessments: ``15\% moisture content detected 3mm beneath paint layer across 2.3m² area, with 85\% confidence of active infiltration based on spectropolarimetric signature correlation with database entry.'' The emergence of initial spectral and polarization vision datasets \citep{jeon2024}, coupled with neural rendering techniques for spectro-polarimetric fields \citep{kim2023}, provides the foundation for transforming property inspection from an experience-based practice to a rigorous, physics-grounded discipline. The realization of this vision depends not primarily on optical hardware refinement but on developing comprehensive spectropolarimetric material databases and robust artificial intelligence frameworks capable of real-time operation on edge computing architectures---ultimately rendering visible the previously imperceptible phenomena critical to accurate property assessment.

\subsubsection{Plenoptic Sensing}

Plenoptic (light field) cameras represent a fundamental reimagining of image capture, recording not just the intensity of light at each pixel but also the direction from which that light arrived. This additional angular information enables remarkable capabilities impossible with conventional imaging: post-capture refocusing to any depth plane, single-shot depth estimation without stereo correspondence, and viewing angle correction for accurate material characterization \citep{liu2024b}. Unlike traditional cameras that collapse the light field by recording only spatial intensity variations, plenoptic systems capture the complete 4D light field \(L(u,v,s,t)\) where \((u,v)\) represents spatial coordinates on the sensor and \((s,t)\) represents the angular direction of incoming rays \citep{levoy1996, ngLightFieldPhotography}. This rich representation could be then extended to more dimensions $L(u,v,s,t,M,\lambda)$, incorporating the Mueller matrix M and spectral information $\lambda$, enabling what might be termed ``physical 3D Gaussian Splatting''---a complete capture of both geometric and material properties. Recent advances demonstrate the enduring importance of light field principles in neural rendering, with multiple approaches combining classical light field theory with modern representations to achieve efficient view synthesis and accurate modeling of view-dependent effects (\citep{sitzmannLightFieldNetworks2021, cao2023, lee2024, wang2025b}).

For spectropolarimetric sensing in indoor inspection scenarios, plenoptic imaging proves particularly valuable by providing the angular information necessary to correct for view-dependent effects on non-Lambertian surfaces (\citep{burkart2015, filip2017},, \citep{lpez2025, solr2025}). Many building materials---polished stone, metallic fixtures, glossy paints---exhibit significant variation in their spectral and polarimetric signatures depending on viewing angle. Traditional approaches would require multiple captures from different positions or complex geometric modeling. Plenoptic cameras solve this elegantly by capturing multiple views simultaneously, enabling accurate material characterization even on surfaces where conventional methods fail \citep{yue2024}. Commercial implementations like Cubert's hyperspectral light field cameras demonstrate the technology's maturity for industrial applications, though significant challenges remain in miniaturizing these systems for drone deployment.

The computational imaging revolution has transformed plenoptic systems from bulky optical assemblies into compact, software-defined sensors suitable for weight-constrained platforms. Traditional light field cameras required either arrays of micro-lenses (as in Lytro's consumer cameras) or multiple synchronized cameras, adding significant weight and mechanical complexity. Modern computational approaches leverage various techniques to capture light field information with minimal hardware. Coded apertures \citep{habuchi2024} modulate the light field during capture, enabling reconstruction from a single sensor. Programmable masks \citep{liang2008} placed at various positions in the optical path create distinct blur patterns that encode depth information. Compressed sensing techniques \citep{gupta2018} exploit the inherent redundancy in natural light fields to reconstruct full 4D information from dramatically undersampled measurements. Most impressively, transversely dispersive metalens arrays achieve 4nm spectral resolution while capturing complete 4D information (3D spatial plus 1D spectral) in a single snapshot using structures just 165$\mu$m thick \citep{yue2024}. Machine learning has proven transformative in enabling reconstruction of full light fields from severely undersampled measurements \citep{cao2024}---what once required 16 synchronized cameras can now be achieved with strategic multiplexing and neural network processing \citep{sitzmannLightFieldNetworks2021}. This fundamental shift from hardware complexity to algorithmic sophistication aligns perfectly with drone payload constraints, where computational resources can be more readily scaled than physical weight capacity.

The commercial landscape for plenoptic technology reveals both significant progress and persistent challenges. While Stanford spinoff Lytro's \citep{ngLightFieldPhotography} consumer photography venture ultimately failed despite innovative technology, industrial and scientific applications have found more sustainable niches. Raytrix dominates the high-end industrial inspection market with their multi-focus plenoptic cameras, but their lightest models still exceed 2 kg---far beyond what small inspection drones can carry. Cubert emerges as the most promising option for drone integration: their ULTRIS S5 hyperspectral light field camera weighs less than 350g and captures 51 spectral bands from 450-850nm, approaching the sub-500g payload limits of advanced inspection drones \citep{cubert2024}. However, even these ``lightweight'' systems push the boundaries of what small autonomous platforms can practically carry while maintaining adequate flight time and maneuverability.

The path forward lies in leveraging metamaterial optics and advanced computational techniques to achieve radical miniaturization. Metasurface folded optics can reduce system thickness to just 0.7mm by manipulating the light path through subwavelength structures rather than conventional lens elements \citep{park2024}. Multi-resonant metasurfaces enable snapshot light field acquisition with minimal training data requirements by encoding angular information directly in the spectral response \citep{zhangRealtimeMachineLearning2024}. These advances, combined with continued improvements in sensor technology and computational algorithms, promise to shrink current kilogram-scale systems to sub-100g implementations within the next 3-5 years---finally making light field capture practical for routine drone deployment.

Computational techniques offer multiple innovative pathways to radical miniaturization of plenoptic systems. Compressive light field imaging exploits the mathematical sparsity of the plenoptic function in various transform domains, enabling accurate reconstruction from measurements that dramatically violate traditional Nyquist sampling requirements. By strategically placing coded masks at specific positions in the optical path, a single sensor can capture multiplexed projections that computational algorithms decode into full 4D light fields---achieving with one sensor what traditionally required dozens \citep{marwah2013}. Fourier slice photography leverages the fundamental mathematical relationship between spatial and angular frequencies, using programmable apertures to capture specific slices of the light field's Fourier transform that can be computationally combined \citep{veeraraghavan2007}. Time-multiplexed approaches trade temporal resolution for spatial-angular sampling density, particularly suitable for static indoor scenes where capture time is less critical than data quality. Most promisingly, learned optical elements co-designed with reconstruction networks achieve optimal information capture for specific tasks---a metasurface designed specifically for material classification might sacrifice general imaging quality to maximize discriminative power at key spectral-angular combinations, achieving better task performance with simpler hardware \citep{baek2023}. These computational approaches can reduce hardware requirements by 10-100x while maintaining or even exceeding traditional performance for targeted applications.

For property inspection applications, the tradeoffs between plenoptic sensing and traditional depth sensors (LiDAR/ToF) require careful consideration based on specific inspection requirements. Plenoptic cameras excel at close-range material characterization where their unique capabilities provide maximum value. Properly calibrated systems achieve sub-millimeter depth precision within 0.05-2.0m distances \citep{monteiro2018}, with angular information enabling several unique capabilities: disambiguating specular from diffuse reflection components crucial for surface finish assessment, measuring surface roughness through angular scattering patterns \citep{sulc2022}, and correcting spectropolarimetric measurements for viewing angle---critical for accurate material classification on non-Lambertian surfaces \citep{pan2020}. Time-of-Flight cameras provide an intermediate solution, achieving high frame rates (up to 160 fps) with moderate accuracy (typically 1\% of measured distance) but are constrained to lower spatial resolutions (typically VGA) \citep{hansard2012}. In practice, completely replacing traditional depth sensors would sacrifice essential long-range navigation capabilities. The ideal solution leverages complementary strengths: plenoptic sensing for detailed close-range inspection where angular information proves crucial, LiDAR for long-range navigation and structural mapping, and ToF for reactive collision avoidance \citep{rezaee2024}. This hybrid approach maximizes inspection capabilities while respecting the severe weight constraints of sub-500g platforms.

The convergence of multiple technological advances points toward transformative possibilities for plenoptic sensing within the next 5-10 years \citep{pan2022, park2024}. Near-term opportunities lie in specialized indoor inspection scenarios where plenoptic cameras' unique capabilities---single-shot depth estimation \citep{hernandezSingleShotMetricDepth2024}, post-capture refocusing for detail examination, and angular-dependent material characterization---justify their current weight premium. Data-driven approaches offer particular promise: using high-quality LiDAR reference data from systems like the Livox Mid-360 or industrial laser scanners \citep{yeshwanth2023, lazarowCubifyAnythingScaling2024} to train neural networks that enhance plenoptic reconstruction quality \citep{labussire2023, wang2020}. This creates a ``computational supervisor'' that improves depth estimation accuracy and reduces artifacts by learning the relationship between plenoptic measurements and ground truth geometry. Collecting comprehensive multimodal datasets combining plenoptic captures with high-precision LiDAR ground truth would accelerate this development \citep{hernandezSingleShotMetricDepth2024}. The ultimate vision involves sub-20g metamaterial sensors providing full light field capture with integrated on-sensor processing \citep{luo2022, fu2024}---perhaps even implementing optical neural networks directly in the metasurface design for zero-latency feature extraction (\citep{hu2024b, heng2024, chamoli2025}). While current technology requires pragmatic compromises between capability and payload, the trajectory is clear: plenoptic sensing will transition from specialized industrial tool to essential component of autonomous inspection systems, fundamentally changing how drones perceive and understand indoor environments through the complete physics of light propagation.

\subsubsection{Thermal Imaging: Hidden Dynamics}

Thermal imaging provides a complementary window into material properties by revealing temperature variations that indicate active physical processes invisible to optical sensors. Unlike hyperspectral imaging that identifies materials through chemical signatures or polarimetric imaging that characterizes surface properties, thermal cameras detect the infrared radiation emitted by all objects above absolute zero, creating temperature maps that reveal moisture evaporation, heat loss, electrical faults, and structural defects through their thermal signatures.

For building inspection applications, thermal imaging offers several critical capabilities. Moisture detection represents one of the most valuable applications, as water's high thermal mass and evaporative cooling create distinctive temperature patterns. Active moisture infiltration appears cooler than surrounding dry materials due to evaporation, while areas of trapped moisture may show different thermal time constants when ambient temperatures change \citep{garrido2019}. This enables inspectors to trace moisture pathways and identify the source of leaks that may be far from visible damage. Energy efficiency audits rely heavily on thermal imaging to identify missing or damaged insulation, which appears as thermal bridges where heat conducts through structural elements \citep{doe2025, sfarra2019}. During heating or cooling cycles, these thermal bridges become clearly visible as temperature anomalies, enabling quantitative assessment of insulation effectiveness \citep{sadhukhan2020}. Electrical systems under load generate heat proportional to resistance and current flow, making thermal cameras invaluable for detecting loose connections, overloaded circuits, and failing components before they cause fires or outages \citep{flir2025}. For insurance applications, thermal imaging can differentiate between old and fresh water damage based on evaporation patterns and thermal mass differences---a critical capability for claim validation and fraud prevention \citep{restoration2025}. For BIM applications, thermal imaging can be used for example to study subsurface concrete delamination \citep{pozzer2025}.

Modern uncooled microbolometer arrays have revolutionized thermal imaging accessibility by eliminating the need for cryogenic cooling while maintaining sufficient sensitivity for building inspection. Sensors like the FLIR Lepton provide 160\texttimes{}120 or 80\texttimes{}60 pixel resolution with thermal sensitivity better than 50mK, all in packages weighing less than 1 gram. While this resolution appears modest compared to visible cameras, it proves adequate for most inspection tasks when combined with higher-resolution visible imagery for context. The integration of thermal data with RGB and depth information enables sophisticated analyses like ThermalGaussian \citep{lu2025}, which creates temperature-aware 3D reconstructions by mapping thermal measurements onto geometric models. This fusion allows inspectors to visualize heat flow through three-dimensional structures, identifying thermal bridges and insulation gaps that would be ambiguous in 2D thermal images alone.

An often-overlooked advantage of thermal imaging is its robustness to environmental conditions that degrade visible and near-infrared sensors. Long-wave infrared radiation (8-14 $\mu$m) experiences significantly less scattering from dust, smoke, and fog compared to shorter wavelengths \citep{li2024b, munir2022}. This makes thermal imaging valuable not only for routine inspection but also for operation in challenging environments such as construction sites with airborne particulates, fire damage assessment through smoke, or emergency response scenarios (\citep{erlenbusch2023, brenner2023, niskanen2024, shin2025}). The dual-use nature of thermal imaging technology---equally valuable for building inspection and search-and-rescue operations---has driven significant investment in miniaturization and integration.

Integrating thermal imaging into multimodal sensing stacks presents unique calibration challenges that must be carefully addressed. Thermal cameras operate in the 8-14 $\mu$m LWIR band, which renders traditional visible-spectrum calibration targets ineffective. Alternative calibration strategies include using heated targets with known emissivity patterns, exploiting sharp thermal gradients at material boundaries, or leveraging the correspondence between geometric edges visible in both thermal and visible imagery (\citep{fu2022, dalirani2023, duAutomaticSpectralCalibration2024, lu2025}). Recent work has demonstrated sub-pixel calibration accuracy between thermal and visible cameras, enabling precise fusion of thermal and spectropolarimetric data for comprehensive material analysis.

The fundamental SWaP (Size, Weight, and Power) constraints of small drones make it impractical to mount all sensing modalities on a single platform while maintaining reasonable flight times. Modern integrated thermal-visible packages have achieved significant miniaturization, with uncooled LWIR sensors reaching 48\% weight reduction compared to previous generations \citep{skydioflir2024}. However, even these optimized packages must remain under 800g when including gimbals and processing hardware. This reality has driven adoption of heterogeneous swarm approaches where specialized drones carry different sensor suites. A ``scout'' drone with lightweight thermal and visible cameras might rapidly survey a property to identify areas of interest, followed by an ``inspector'' drone carrying heavier hyperspectral or polarimetric sensors for detailed analysis of anomalies. This division of labor yields superior results compared to compromising every sensing capability in pursuit of an all-in-one platform that violates fundamental physics and battery constraints.

The dual-use applications of thermal-spectropolarimetric fusion extend beyond building inspection. The same algorithms that detect moisture infiltration in buildings can identify camouflaged objects in security applications, as both involve detecting material anomalies against complex backgrounds. Real-time calibration correction indoor inspection by compensating for temperature drift during extended flights. This convergence of civilian and security applications accelerates technology development and reduces costs through larger production volumes, ultimately benefiting property inspection applications with more capable and affordable thermal imaging solutions.

\subsubsection{Emerging Frontiers: Event Cameras and Beyond}

Beyond the established modalities of hyperspectral, polarimetric, and thermal imaging, several emerging sensing technologies promise to address fundamental limitations of frame-based vision systems and passive optical sensing. These nascent technologies, while still largely in research phases, offer unique capabilities that could transform autonomous inspection once technical and practical challenges are overcome.

Event cameras represent perhaps the most mature of these emerging technologies, operating on a fundamentally different principle than traditional frame-based cameras. Rather than capturing full images at fixed intervals, event cameras asynchronously report pixel-level brightness changes the moment they occur, achieving microsecond temporal resolution with minimal power consumption. Each pixel operates independently, generating events only when the logarithmic intensity change exceeds a threshold, resulting in a sparse data stream that naturally compresses static scenes while capturing high-speed dynamics (\citep{falanga2019, falanga2020, gehrig2024}). For drone navigation in cluttered indoor environments, these characteristics prove invaluable---event cameras maintain performance in high-dynamic-range scenarios where conventional cameras suffer from motion blur or exposure problems. Racing drones equipped with event cameras have demonstrated superhuman performance in navigating complex environments at high speed, suggesting similar advantages for inspection drones maneuvering through confined spaces.

Despite their advantages, event-based perception remains underutilized in real estate and indoor robotics applications due to several practical challenges. The asynchronous, sparse nature of event data requires fundamentally different processing algorithms than traditional computer vision pipelines developed over decades for frame-based imagery. Events encode temporal contrast rather than absolute intensity, making simple tasks like object recognition non-trivial. High-frequency noise, particularly in low-light conditions, can overwhelm the signal, while the nonlinear logarithmic encoding complicates radiometric measurements. Perhaps most critically, the lack of standardized datasets and evaluation metrics for indoor inspection tasks has hindered algorithm development. Addressing these challenges, the recently introduced SENPI framework provides a differentiable event camera simulator for PyTorch, enabling researchers to evaluate noise models, contrast thresholds, and fusion architectures using synthetic and real data interchangeably \citep{greene2025}. This tool promises to accelerate development of event-based perception algorithms specifically tailored for property inspection applications.

Millimeter-wave (mmWave) radar offers unique capabilities for penetrative sensing, maintaining robust performance in conditions that completely defeat optical sensors---smoke, dust, complete darkness, and even through certain building materials. Unlike optical sensors that capture surface properties, mmWave radar can potentially detect subsurface features like pipes, wiring, and structural elements hidden within walls (\citep{lu2020, chen2024b, brescia2024}). This capability would prove invaluable for comprehensive structural assessment, enabling inspectors to verify as-built conditions against architectural plans or detect unauthorized modifications. Current mmWave sensors produce relatively sparse point clouds compared to LiDAR, but algorithmic advances in super-resolution and machine learning-based enhancement hint at near-term improvements (\citep{amin2014, pham2022, murakami2024}). Recent developments have been particularly promising: phase error correction techniques for handheld mmWave devices have achieved sub-wavelength precision in through-wall imaging \citep{chen2023}, while sparse sampling strategies have reduced antenna array costs by an order of magnitude without sacrificing resolution \citep{bian2024}. These advances are rapidly transitioning mmWave sensing from laboratory curiosity to practical inspection tool, particularly for scenarios where knowing what lies behind surfaces is critical for accurate assessment.

Raman spectroscopy occupies a unique position in the sensing landscape by providing molecular-level chemical identification through vibrational spectra. While hyperspectral imaging infers chemical composition from electronic transitions, Raman spectroscopy directly probes molecular bonds, offering orthogonal and often more specific material identification (\citep{jehlika2022, ilchenko2024, guo2025}). This specificity proves particularly valuable for identifying hazardous materials, authenticating high-value components, or detecting chemical contamination that might be ambiguous in hyperspectral data. Recent advances have dramatically improved the practicality of Raman systems for field deployment. Standoff Raman systems can now perform chemical analysis at distances up to 15 meters under ambient light conditions, using advanced filtering and time-gating to reject fluorescence and background illumination \citep{li2019, li2024c}. For drone integration, the challenge lies in miniaturizing laser sources and spectrometers while maintaining sufficient sensitivity. Photonic integrated circuits and MEMS-based spectrometers show promise for achieving the required size reductions, though current prototypes still exceed practical weight limits for small drones.

Several other sensing modalities remain in earlier stages of development but offer intriguing possibilities for future integration. Acoustic sensing could enable void detection and material density assessment through ultrasonic imaging, potentially identifying delamination in composite materials or voids in concrete structures \citep{ju2023}. Through-wall radar imaging at lower frequencies than mmWave could provide deeper penetration for locating structural elements, plumbing, and electrical systems, though resolution decreases with wavelength \citep{nkwari2017}. Airborne chemical sensors targeting specific volatile organic compounds (VOCs) could detect mold growth, gas leaks, or off-gassing from building materials, providing early warning of air quality issues \citep{li2024d, tarvo2024}. More speculative approaches include exploiting the effect of indoor airflow on drone flight dynamics to map 3D ventilation patterns \citep{xia2023}, or using precision light meters to verify lighting quality against design specifications and standards (\citep{pizg2022, tarvo2024, song2021, hanqqa2022}). While outdoor applications have demonstrated drone-based light pollution mapping \citep{bobkowska2024}, indoor lighting assessment presents unique challenges due to complex reflection patterns and multiple source types.

These emerging sensing modalities collectively support a modular architecture where task-specific sensor payloads can be dynamically deployed based on inspection objectives \citep{srensen2017}. Rather than attempting to integrate every possible sensor onto a single platform---an approach doomed by physics and practicality---future inspection systems will likely employ heterogeneous swarms where different drones carry complementary sensor suites. This flexibility enables optimization for specific inspection scenarios while managing the fundamental tradeoffs between sensing capability, platform agility, and operational endurance. As these technologies mature from laboratory demonstrations to field-ready systems, they promise to extend autonomous inspection capabilities far beyond what current visual and thermal sensors can achieve, ultimately enabling truly comprehensive understanding of building conditions through multiple physical sensing modalities.

\subsection{The Computational Imaging Revolution: Sensor-AI Co-Design}

The future of drone-deployable advanced sensing doesn't lie in shrinking traditional instruments to fit on small platforms. Instead, we must fundamentally reimagine optical system design itself, integrating even large language models into the process \citep{kim2025}. This reimagining takes the form of computational imaging---a paradigm where optical hardware and reconstruction algorithms evolve together as an integrated system, each component designed with the other in mind (\citep{wang2025c, frch2025, chen2025c}).

\subsubsection{Metasurfaces: Flat Optics for Radical Miniaturization}

The cornerstone of this revolution is the metasurface---an ultrathin array of subwavelength nanostructures that manipulates light in ways that would require entire optical benches using conventional components. These surfaces, thinner than a human hair, control the phase, amplitude, and polarization of light at each point across their area, essentially programming the behavior of photons as they pass through.

Consider what this means for drone-based sensing. A single multi-wavelength metasurface chip can capture hyperspectral information using remarkably sparse training data---just 18 data points in one demonstration \citep{lin2023}. Chiral metasurfaces eliminate the need for rotating polarizers entirely, enabling full-Stokes polarimetry in a static configuration \citep{basiri2019, deng2024a}. Perhaps most intriguingly, these capabilities can be combined---researchers have demonstrated single metasurfaces that simultaneously encode both spectral and polarimetric information \citep{zhangRealtimeMachineLearning2024}.

\subsubsection{End-to-End Learning: Task-Specific Optimization}

Traditional optical design follows a sequential philosophy: create the best possible image, then extract information from it. Computational imaging inverts this approach entirely. In end-to-end learned systems, the entire pipeline---from the metasurface's nanostructure pattern to the neural network's weights---is jointly optimized for a specific sensing task \citep{arya2024}.

Take moisture detection as an example. Rather than capturing full hyperspectral datacubes and then searching for water's spectral signatures, the metasurface itself could be optimized to maximize contrast specifically at water's absorption bands. The accompanying neural network learns to map these tailored optical encodings directly to moisture probability maps. This task-specific co-design achieves superior performance with dramatically simpler hardware \citep{roques-carmesComputationalMetaopticsImaging2024, hu2024a}.

Making this possible are differentiable optics simulators that allow gradient information to flow backward through the entire imaging chain (\citep{dekoning2023, hoDifferentiableWaveOptics2024, tseng2021}). Physics-informed neural networks (PINNs) ensure that optimized designs respect fundamental constraints---Maxwell's equations govern electromagnetic propagation while fabrication tolerances limit achievable feature sizes \citep{khoram2020}. These frameworks produce systems that seem to defy traditional optical trade-offs \citep{jenkins2021, schubert2022}.

Beyond fixed designs, electrically tunable metasurfaces introduce unprecedented adaptability. Materials like graphene \citep{abadal2019}, liquid crystals \citep{zhang2025e}, and phase-change materials \citep{shalaginov2021} enable real-time reconfiguration of optical properties. For drone inspection, this means a single sensor could switch between detection modes on demand---from identifying water damage (optimizing for 1450nm absorption) to detecting mold growth (targeting specific spectral signatures) to mapping thermal bridges (enhancing temperature gradient sensitivity), all through electrical control signals.

\subsubsection{Practical Challenges and Solutions}

Despite its elegance, computational imaging faces significant real-world hurdles. The simulation-to-reality gap remains substantial---designs that perform perfectly in silico may fail catastrophically when fabricated due to manufacturing variations \citep{daiToleranceAwareDeepOptics2025}. Environmental factors like temperature fluctuations and mechanical vibrations can degrade carefully optimized performance \citep{caiHighReliabilityDamage2023, vujii2019}. Perhaps most challenging, these learning-based systems typically require extensive training datasets that may not exist for specialized inspection scenarios \citep{leeUnlockingCapabilitiesExplainable2024a}.

The field is developing sophisticated solutions to these challenges. Fabrication-aware optimization incorporates manufacturing constraints directly into the design process, ensuring that theoretical performance translates to physical devices. Self-calibrating architectures use redundant optical encodings to detect and compensate for system drift in real-time \citep{jiang2019}. Few-shot learning techniques and synthetic data generation reduce the burden of training data collection \citep{sattar2019, liu2024c}. Looking ahead, adaptive systems that continue learning during deployment may provide the ultimate solution \citep{waller2017}.

\subsection{The Calibration Challenge: The Hidden Hurdle}

While sensor miniaturization captures attention, inter-sensor calibration remains the unsung challenge of multimodal sensing. Even a perfect stack of sensors becomes worthless if their data streams cannot be precisely aligned in space and time. This alignment requires determining exact 6-DOF extrinsic parameters---the relative positions and orientations between all sensors in the system \citep{qiuExternalMultimodalImaging2023}.

The challenge intensifies when sensors operate on fundamentally different physical principles. How do you align hyperspectral data revealing chemical composition with polarimetric measurements probing surface physics and thermal images mapping temperature distributions? Traditional calibration targets like checkerboards fail when sensors cannot perceive common features \citep{persicCalibrationHeterogeneousSensor2018}.

Contemporary solutions exploit motion as a universal reference frame. As the sensor suite moves through space, each modality observes the world transforming according to the same underlying motion, providing a common signal for alignment (\citep{hayounPhysicsSemanticInformed2024, taylorMotionbasedCalibrationMultimodal2015, lv2025}). Deep learning methods discover subtle cross-modal correlations, achieving sub-centimeter alignment accuracy even between sensors as disparate as LiDAR and polarimetric cameras \citep{zhao2021}.

Most critically, online calibration algorithms continuously refine these parameters during operation, compensating for vibrations and thermal expansion that would otherwise degrade alignment \citep{liuNovelMotionbasedOnline2022, pengAutomaticMiscalibrationDetection2024}. This sophisticated calibration infrastructure represents a significant technical barrier---competitors might acquire identical sensors, but without the complex algorithms to fuse their outputs coherently, they possess only a collection of misaligned measurements rather than actionable insights.

\subsection{Toward Multimodal Spatial Intelligence}

The emergence of neural rendering techniques, particularly 3D Gaussian Splatting (3DGS), offers an unprecedented opportunity to unify geometric reconstruction with rich physical sensing. This convergence promises to transform indoor inspection drones from passive cameras into active investigators that reveal the hidden physics of built environments.

3DGS's power lies in its elegant representation---each scene is modeled as a collection of 3D Gaussian primitives, defined by position, covariance, and associated attributes. This simple yet flexible framework naturally accommodates heterogeneous sensor data, spawning two distinct integration philosophies.

\begin{figure*}
    \centering
    \includegraphics[width=1\textwidth]{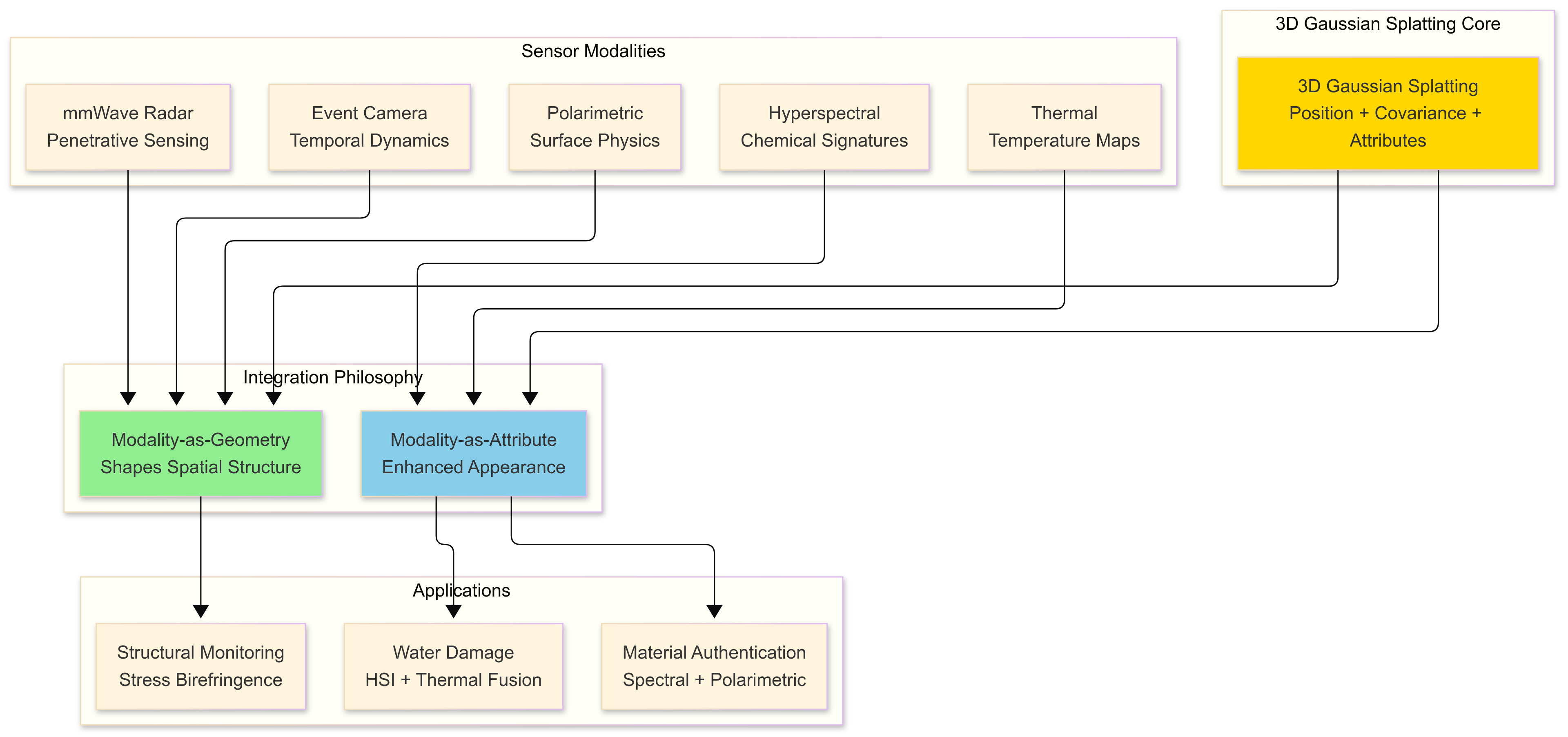}
    \caption{Two approaches to integrating physics-aware sensing with 3D Gaussian Splatting: Modality-as-Attribute treats sensor data as enhanced appearance (like RGB color)—hyperspectral and thermal data become Gaussian attributes, computationally simple but missing geometric insights. Modality-as-Geometry reshapes spatial structure—polarimetry guides surface normals, radar/LiDAR provide geometric constraints, and event cameras add temporal dynamics. Applications include: water damage mapping by fusing 1450nm hyperspectral absorption with thermal cooling signatures; material authentication combining spectral and polarimetric analysis; and structural monitoring via stress birefringence detection before visible failure}
    \label{fig:multimodal-sensing}
\end{figure*}

The ``Modality-as-Attribute'' approach treats additional sensing as enhanced appearance information (see \autoref{fig:multimodal-sensing}). Just as traditional 3DGS assigns RGB colors to each Gaussian, multimodal variants assign hyperspectral signatures or thermal readings instead. HyperGS \citep{thirgood2024} tackles the curse of dimensionality by performing view synthesis in a learned latent space, compressing hundreds of spectral bands into manageable representations. ThermalGaussian \citep{lu2025} introduces multimodal regularization to prevent the system from overfitting to any single sensing modality.

MS-Splatting pushes further, introducing a calibration-free framework where thermal and multi-spectral data are encoded through compact neural embeddings, enabling joint spectral rendering with improved cross-band fidelity \citep{meyer2025}. SpectralGaussians adds semantic awareness, using per-spectrum reflectance and lighting estimates to enable applications like spectral inpainting and cross-modal style transfer \citep{sinha2025}.

The alternative ``Modality-as-Geometry'' philosophy represents a deeper reimagining where sensor data shapes the spatial structure itself  (see \autoref{fig:multimodal-sensing}). GNeRP \citep{li2024e} exemplifies this approach by using polarization measurements to guide surface normal estimation---the Gaussian field encodes geometric properties rather than just appearance. LiDAR integration methods like TCLC-GS \citep{lei2024} and LiDAR-3DGS \citep{lim2024} provide direct geometric supervision for Gaussian placement and scaling. Time-of-flight arrays offer similar geometric constraints \citep{conti2025}. This distinction becomes crucial when reconstructing challenging materials---transparent surfaces or specular reflectors where photometric cues fail but polarimetric or range measurements provide reliable geometric guidance.

Each sensing modality brings unique integration challenges. Hyperspectral imaging's high dimensionality demands sophisticated compression through learned encodings \citep{thirgood2024}. Thermal imaging requires physics-based modeling of heat diffusion and atmospheric effects, as demonstrated by Thermal3D-GS \citep{chen2024d} which incorporates thermal conduction modules to sharpen blurry thermal boundaries. Event cameras introduce temporal dynamics that fundamentally alter the reconstruction pipeline, requiring accumulation strategies seen in E2GS \citep{yura2024} and incremental updates in IncEventGS \citep{zhang2024a}.

Radar and mmWave sensors present particularly interesting challenges due to their sparse, noisy returns contaminated by multipath reflections and receiver saturation effects. RadarSplat \citep{kung2025} addresses these by explicitly modeling radar-specific artifacts within the 3DGS framework. Recent breakthroughs are transforming mmWave radar from a coarse proximity sensor into a viable 3D imaging modality. RFconstruct \citep{hussein2025} demonstrates how fusing orthogonal radar views with odometry-aware temporal aggregation can reconstruct detailed 3D shapes without prior object knowledge. Meanwhile, range image-conditioned diffusion models \citep{wu2025a} leverage the natural projection geometry of radar to enhance resolution and suppress noise, producing LiDAR-like point clouds from commodity radar hardware.

The implications for autonomous inspection extend far beyond simple sensor fusion. Consider water damage assessment: multimodal 3DGS enables comprehensive moisture mapping where hyperspectral absorption signatures at specific wavelengths combine with thermal signatures of evaporative cooling to reveal not just moisture presence but penetration depth and flow patterns. Material authentication becomes virtually foolproof when spectral signatures revealing chemical composition integrate with polarimetric patterns exposing surface microstructure. Structural monitoring benefits from observing stress-induced birefringence alongside thermal expansion patterns, enabling early failure detection.

MM3DGS SLAM \citep{luo2024a} demonstrates the power of tight sensor coupling within the 3DGS framework, showing substantial improvements in both tracking accuracy and rendering quality when vision, depth, and inertial measurements are jointly optimized. Agricultural applications showcase similar benefits \citep{sahoo2025}, where multimodal integration enables precise crop monitoring.

The future points toward adaptive systems that dynamically adjust sensing strategies based on scene understanding \citep{varotto2021}. The unified representation enables closed-loop perception---initial reconstructions guide subsequent data acquisition, perhaps triggering detailed hyperspectral scanning upon detecting thermal anomalies or initiating polarimetric analysis where material ambiguities arise. This active perception paradigm \citep{placed2023} transforms drones from predetermined flight paths into intelligent investigators pursuing hypotheses \citep{hickling2025}. Systems like DroneSplat \citep{tang2025b} and DRAGON \citep{chen2024e} already demonstrate robust reconstruction from challenging drone footage, foreshadowing seamlessly integrated multimodal systems.

Significant challenges remain before this vision becomes reality. Calibration between fundamentally different sensors requires sophisticated techniques beyond traditional checkerboard patterns (\citep{herau2024, zhou2025b, jung2025, qu2021, qiuExternalMultimodalImaging2023}). Processing multiple high-dimensional data streams taxes even modern edge computing platforms. Storage and transmission of spectropolarimetric datacubes generates orders of magnitude more data than RGB video.

Perhaps most critically, the lack of comprehensive multimodal indoor datasets hampers progress \citep{cao2024a}. While outdoor datasets serve autonomous driving research \citep{alibeigi2023, zheng2025}, indoor inspection's unique challenges---variable illumination, confined spaces, diverse materials---remain underrepresented (\citep{lee2021, kaveti2023, bamdad2024}). Progress requires coordinated efforts across the community \citep{udandarao2024}.

Future inspection platforms will likely employ hierarchical processing \citep{hughes2024}, where lightweight modalities guide computationally intensive sensors. Federated learning could enable drone fleets to collaboratively build material signature databases without transmitting raw data \citep{khan2024}. Physics-informed constraints will ensure reconstructions respect thermodynamic principles \citep{zhang2022, kamali2024}.

The emergence of foundation models for 3D understanding (\citep{zuo2024, jiang2025b, chen2025d}) suggests future systems might leverage pre-trained representations that generalize across modalities and tasks. As these technologies mature, we anticipate a fundamental transformation in robotic perception---from passive observation to active understanding, from appearance capture to physics discovery, from isolated measurements to integrated spatial intelligence (\citep{bajcsy2017, firoozi2024, han2025a, ruan2025}). This evolution extends beyond property inspection to medical imaging, quality control, and environmental monitoring---anywhere understanding hidden physics matters as much as surface appearance.

\section{Intelligent Autonomy: From Passive Scanning to Active Understanding}

The fundamental challenge in autonomous indoor mapping extends beyond simple navigation through buildings---it requires intelligent determination of optimal viewpoints to construct comprehensive 3D models with maximum efficiency. This challenge, termed active reconstruction, next-best-view selection, or active image selection depending on the research community (\citep{placed2023, chen2024f, wang2025d, polyzos2025}), transforms drones from passive data collectors following predetermined waypoints into intelligent agents reasoning about information value. For property assessment applications where every minute on-site directly impacts operational costs, the distinction between passive and active approaches can determine commercial viability of drone-based inspection services.

\subsection{Understanding the Mapping Pipeline: Localization, Mapping, and Reconstruction}

Before examining active reconstruction strategies, it is essential to understand three interconnected but distinct computational problems that any autonomous indoor drone must solve simultaneously:

Localization addresses the fundamental question ``Where am I?'' In the absence of GPS signals indoors, drones must continuously estimate their position and orientation (pose) relative to a starting reference frame. This is typically achieved through Visual-Inertial Odometry (VIO), which fuses visual features extracted from camera imagery with high-frequency measurements from Inertial Measurement Units \cite{famili2022, kinnari2024}. Modern VIO algorithms achieve centimeter-level accuracy essential for ensuring complete coverage while avoiding collisions in cluttered indoor environments \citep{khne2024, yuan2025a}.

Mapping addresses ``What's around me?'' by constructing spatial representations of the environment as the drone explores. These representations typically take the form of point clouds, occupancy grids, or more sophisticated data structures that serve dual purposes: enabling safe navigation by identifying obstacles and free space, and providing the geometric scaffold upon which detailed models are built \citep{samavedula2025, ren2024}. For property assessment, this is analogous to sketching accurate floor plans and identifying all rooms requiring inspection \citep{wang2025e}.

3D Reconstruction addresses ``What does it look like?'' by creating photorealistic, textured models that capture visual details, material properties, and surface conditions \citep{su2024, gan2025}. Modern methods leveraging 3D Gaussian Splatting (3DGS) achieve real-time reconstruction quality previously impossible with traditional approaches \citep{li2024f, polyzos2025}. This capability transforms property assessment from knowing a kitchen exists to being able to evaluate cabinet conditions, countertop materials, and appliance states through detailed visual inspection.

The convergence of these three problems through modern SLAM (Simultaneous Localization and Mapping) systems represents a significant advancement in robotic perception. Traditional SLAM focused primarily on sparse geometric representations sufficient for navigation, but recent integration with 3D Gaussian Splatting enables simultaneous tracking and photorealistic reconstruction. Systems like MonoGS++ \citep{li2025e} and Gaussian Splatting SLAM \citep{matsuki2024} demonstrate real-time performance on single GPUs, creating detailed models while maintaining accurate localization.

This convergence fundamentally changes the nature of indoor mapping. Rather than separate pipelines for navigation and visualization, modern systems create unified representations serving both purposes. The integration of depth measurements proves particularly crucial, enabling metric scale recovery essential for property assessment where precise room dimensions and feature measurements are required. Advanced multimodal SLAM systems including MM3DGS SLAM \citep{luo2024a}, Gaussian-LIC2 \citep{li2025c}, and LVI-GS \citep{zhang2025b} demonstrate how tight coupling of visual, depth, and inertial measurements within the 3DGS framework achieves superior tracking accuracy and reconstruction quality.

The distinction between SLAM and pure 3D reconstruction has effectively dissolved with these advances. While traditional SLAM prioritized localization accuracy with mapping as a byproduct \citep{tosiHowNeRFs3D2025}, systems like RTG-SLAM achieve comprehensive scene understanding with approximately twice the speed and half the memory consumption compared to NeRF-based approaches \citep{peng2024}. Semantic SLAM approaches exemplified by SGS-SLAM \citep{li2024g} embed semantic information directly into the 3D Gaussian representation, enabling systems to not only map geometry but understand and classify different building elements---distinguishing kitchens from bathrooms, identifying structural features, and recognizing standard fixtures.

This semantic understanding, combined with metric scale recovery through dual-pixel sensors \citep{ashidaResolvingScaleAmbiguity2024} or depth integration \citep{qin2025, li2025c}, transforms SLAM into a comprehensive scene understanding system suitable for real-time property documentation. Performance benchmarks demonstrate photorealistic rendering at real-time frame rates on consumer hardware such as NVIDIA RTX 4090 GPUs \citep{ha2024}, making these capabilities practical for field deployment rather than merely laboratory demonstrations.

\subsection{The Core Challenge: Uncertainty and Information Gain}

Active reconstruction fundamentally revolves around managing and strategically reducing uncertainty in 3D models. Every reconstruction contains inherent uncertainty arising from multiple sources: sensor measurement noise, occlusions preventing complete observation, limited viewing angle diversity, and accumulated pose estimation errors. The key insight driving active reconstruction is that viewpoints possess dramatically different information values---some observations substantially reduce model uncertainty while others provide largely redundant information \citep{klasson2024}.

This creates a complex optimization problem for autonomous drones operating under severe battery constraints (ranging from 10 minutes for ultra-lightweight platforms like Crazyflie 2.1 Brushless to 55 minutes for larger systems like ModalAI Starling 2 Max): which sequence of viewpoints will produce the highest-quality reconstruction within available flight time? This optimization extends beyond simple geometric coverage to strategic uncertainty reduction in regions most critical for the inspection task.

Modern active reconstruction systems make uncertainty quantification explicit rather than implicit. ActiveGS \citep{jin2025} maintains per-primitive confidence scores for each 3D Gaussian in the reconstruction, enabling targeted acquisition of views that specifically address high-uncertainty regions. NARUTO \citep{feng2024} takes this further by learning continuous uncertainty fields over entire scenes, providing smooth predictions of reconstruction quality even in unobserved areas. These uncertainty estimates guide exploration strategies that ensure ambiguous surfaces receive additional observations from informative angles until quality thresholds are satisfied.

For insurance and appraisal applications, this uncertainty-aware approach provides critical capabilities. It enables systems to distinguish between reconstructions that might have missed subtle defects due to insufficient observation versus those guaranteeing comprehensive inspection to specified resolution standards. This distinction proves essential for liability assessment, warranty claims, and regulatory compliance where inspection completeness must be verifiable rather than assumed.

\subsection{Task-Aware Exploration: From Generic to Application-Specific}

The evolution from generic exploration strategies to task-specific, application-aware approaches represents an advancement in autonomous indoor mapping. Modern systems dynamically adapt their exploration strategies, reward functions, and quality metrics based on specific inspection requirements rather than treating all environments uniformly. This adaptive capability proves particularly crucial for property assessment where different use cases demand fundamentally different tradeoffs between speed, coverage completeness, and reconstruction fidelity (\autoref{fig:multimodal-sensing}).

The evolution from generic exploration strategies to task-specific, application-aware approaches represents an advancement in autonomous indoor mapping \citep{muoz2016, eldemiry2022}. Modern systems dynamically adapt their exploration strategies, reward functions, and quality metrics based on specific inspection requirements rather than treating all environments uniformly \citep{eldemiry2022}. This adaptive capability proves particularly crucial for property assessment where different use cases demand fundamentally different tradeoffs between speed, coverage completeness, and reconstruction fidelity \citep{raj2024, cai2023}.

Sophisticated active reconstruction systems implement multiple operational modes tailored to specific applications:

Fast Floorplan Mode prioritizes rapid frontier exploration to establish room boundaries, connectivity, and gross square footage calculations. Research by \cite{wang2025e} and \citep{song2025} demonstrates optimization strategies specifically targeting floorplan topology understanding rather than photorealistic quality, enabling complete building layout capture in minimal time.

UAD 3.6 Appraisal Mode shifts to methodical, high-fidelity scanning of specific elements mandated by appraisal standards. The reward function prioritizes acquisition of multiple viewpoints for kitchens, bathrooms, HVAC systems, and indicators of damage or exceptional wear. The VISTA framework \citep{nagamiVISTAOpenVocabularyTaskRelevant2025} exemplifies how Large Language Models can parse natural language inspection requirements and generate semantic search queries that guide exploration toward task-relevant features.

Forensic Inspection Mode maximizes data quality over speed for insurance claims or high-value property documentation. Systems may dedicate entire flight segments to capturing detailed multi-angle views of suspected water damage, structural cracks, or material deterioration, using uncertainty metrics to ensure no potentially important detail escapes documentation.

This task-aware capability transforms robotic inspection from one-size-fits-all scanning to intelligent systems that dynamically balance time and quality based on specific business requirements and regulatory constraints.

\begin{figure}
    \centering
    \includegraphics[width=1\linewidth]{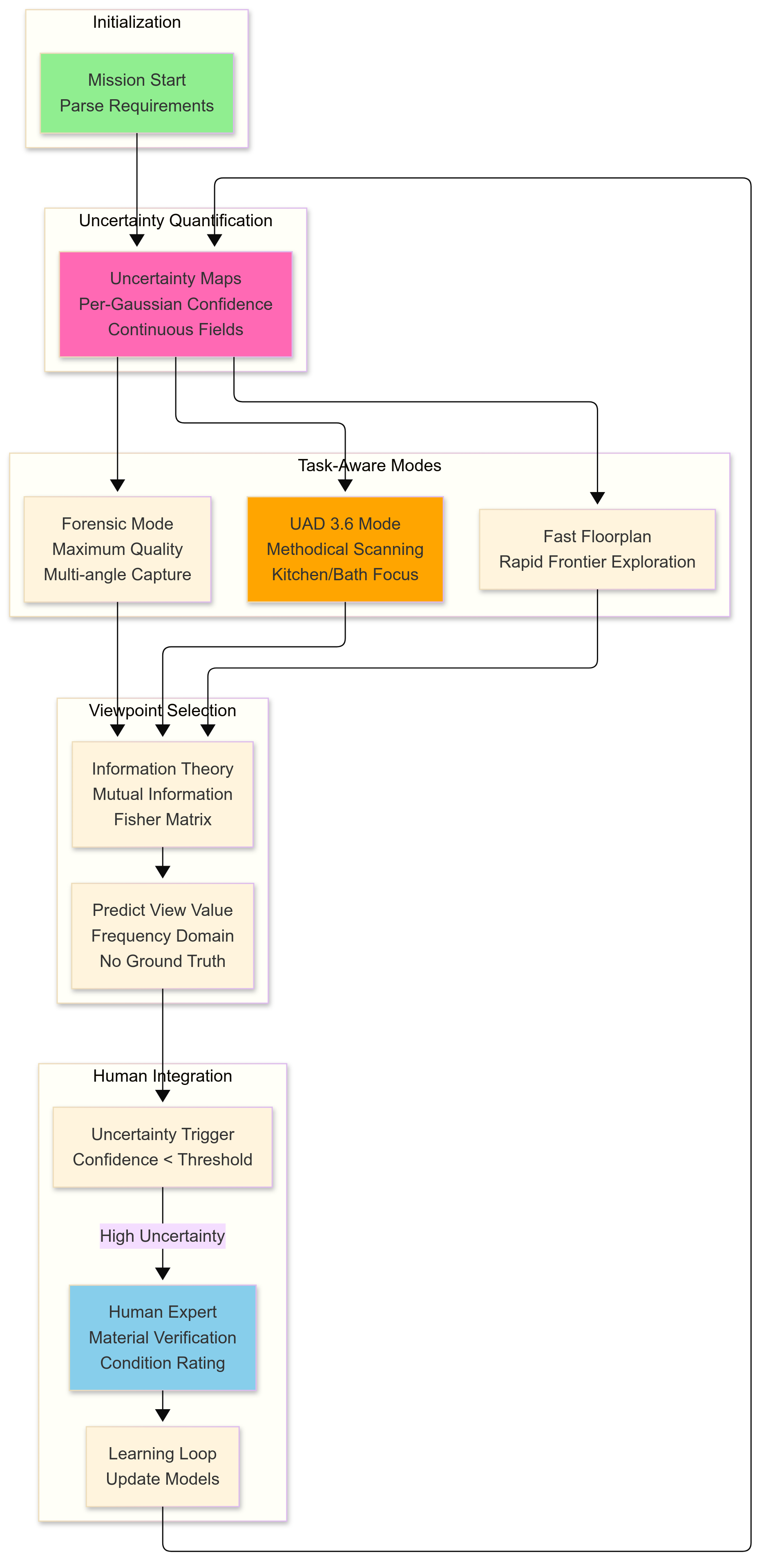}
    \caption{Intelligent viewpoint selection transforms drones from passive scanners to uncertainty-aware active agents. Per-Gaussian confidence scores drive three task-specific modes: Fast Floorplan (rapid topology mapping), UAD 3.6 (methodical kitchen/bath/HVAC scanning for appraisals), and Forensic (multi-angle insurance documentation). Information-theoretic selection using mutual information (GauSS-MI, \cite{xie2025}) and Fisher Matrix analysis (AG-SLAM, \cite{jiang2024a}) predicts uncertainty-reducing views. When confidence drops below thresholds, human-in-the-loop triggers for material verification or condition ratings (C1-C6), generating training data that progressively reduces intervention needs—evolving from constant oversight to strategic consultation.}
    \label{fig:active-recon}
\end{figure}

\subsubsection{Human-in-the-Loop Intelligence for Task-Aware Reconstruction}

While full autonomy remains an aspirational goal for complex indoor environments, the reality of professional property assessment necessitates thoughtful integration of human expertise \citep{langerman2025}. The challenge lies not in whether to include human operators but in designing systems that minimize intervention requirements while maximizing the value of human input when provided. Recent advances in human-drone interaction research (\citep{funk2019, tezza2019, faulkner2024}) reveal that effective human-robot collaboration requires careful consideration of cognitive load distribution, interface design principles, and dynamic task allocation strategies.

The integration of uncertainty-aware human-in-the-loop (HITL) capabilities represents a particularly promising direction. \citep{manderson2025} demonstrate how diffusion policy-based systems can proactively seek human assistance only when autonomous capabilities reach their limits, using uncertainty metrics to determine when operator intervention would provide maximum value. This approach fundamentally transforms the human role from constant system monitor to on-demand expert consultant, dramatically reducing labor requirements while maintaining operational reliability.

For UAD 3.6 appraisal tasks, intelligent HITL integration manifests in several concrete operational scenarios:

Ambiguous Material Classification occurs when spectropolarimetric signatures fall between established material classes---for instance, engineered wood products that share characteristics with both genuine hardwood and laminate flooring. Rather than forcing potentially incorrect autonomous decisions, the system presents high-resolution imagery alongside spectral data for human verification. This intervention both resolves immediate ambiguity and provides labeled training data improving future autonomous classification performance.

Condition Rating Boundaries present challenges as UAD 3.6 requires subjective condition ratings (C1-C6) that often fall near decision thresholds. When autonomous confidence in rating assignment drops below acceptable thresholds---perhaps due to conflicting indicators such as new appliances in kitchens with dated cabinetry---the system presents comprehensive evidence for human determination. This preserves regulatory requirements for appraiser judgment while providing quantitative support for decisions.

Damage Extent Verification scenarios arise when thermal and moisture signatures indicate potential water intrusion, but distinguishing between active leaks requiring immediate attention and historical damage that has been properly remediated requires contextual understanding beyond current autonomous capabilities. Human expertise informed by broader property context enables accurate classification critical for insurance claims and repair recommendations.

Priority Area Identification leverages human domain knowledge to guide inspection focus. Experienced appraisers might identify subtle indicators during initial walkthroughs---slight floor sagging suggesting structural issues, or discoloration patterns indicating potential mold---that warrant detailed investigation. The autonomous system adapts its exploration strategy to ensure comprehensive documentation of these areas while maintaining overall efficiency.

The bidirectional learning capability of modern HITL systems creates virtuous improvement cycles. Human interventions not only resolve immediate operational needs but generate valuable training data for system enhancement. Manderson et al.~(2025) demonstrate that uncertainty-based intervention mechanisms serve dual purposes: maintaining operational reliability while efficiently collecting precisely the data needed for policy refinement. This creates systems that progressively require less human oversight through deployment experience.

Interface design for property assessment applications must accommodate the expertise profiles of actual operators. Unlike research environments staffed by robotics specialists, commercial deployment involves appraisers, inspectors, and insurance adjusters possessing deep domain knowledge but limited robotics experience. Modern approaches leverage natural language interfaces enabling guidance in professional terminology, augmented reality visualizations overlaying inspection data on live imagery, and semantic scene understanding allowing task specification in familiar terms rather than low-level robot commands.

The economic implications of intelligent HITL systems prove substantial for commercial viability. Reducing human intervention requirements by an order of magnitude---from constant flight monitoring to occasional expert consultation---fundamentally changes operational economics. Single operators can potentially oversee multiple simultaneous inspections across different properties, intervening only when autonomous systems encounter scenarios requiring human judgment. This multiplicative effect on human productivity, combined with quality assurance benefits of selective expert oversight, positions HITL task-aware reconstruction as a practical near-term solution bridging current capabilities with future full autonomy.

\subsection{Hierarchical Reinforcement Learning: The Strategic Brain}

The complexity of indoor active reconstruction, with its long planning horizons, sparse rewards, and multi-scale decision requirements, makes it an ideal application domain for Hierarchical Reinforcement Learning (HRL) \citep{wang2024e}. Recent research demonstrates that HRL excels precisely in environments exhibiting these characteristics, decomposing intractable monolithic problems into manageable hierarchical components \citep{nachum2018, wang2024f}.

HRL architectures for indoor exploration typically implement multiple levels of abstraction:

\begin{itemize}
    \item The High-level ``Strategist'' operates at semantic and room-scale granularity, reasoning about inspection objectives and coverage requirements. This policy might determine sequencing decisions such as ``inspect the master bathroom next'' or ``ensure complete documentation of all HVAC equipment.'' Operating on timescales of minutes, it considers mission-level objectives, regulatory requirements, and resource constraints in planning inspection sequences.
    \item The Low-level ``Pilot'' handles the complex dynamics of physical flight execution, managing trajectory planning and obstacle avoidance while maintaining stable flight. Operating on millisecond timescales, this policy must handle challenging aerodynamic effects near walls and ceilings, dynamic obstacle avoidance, and precise positioning for sensor data acquisition.
\end{itemize}

This hierarchical decomposition offers several critical advantages over monolithic reinforcement learning approaches:

\begin{itemize}
    \item Tractable Learning emerges from problem decomposition where each level learns substantially simpler tasks. The Strategist need not understand rotor dynamics or aerodynamic effects, while the Pilot requires no knowledge of inspection completeness criteria or appraisal standards. This separation enables more efficient learning with better generalization.
    \item Interpretability allows operators to understand and potentially override high-level strategic decisions (``inspect kitchen before bathroom'') while trusting low-level safety behaviors to maintain flight stability. This transparency proves essential for commercial deployment where operators must maintain situational awareness and regulatory compliance.
    \item Transfer Learning capabilities enable component reuse across different environments. A Pilot policy trained in one building transfers effectively to another with similar construction, while only the Strategist requires environment-specific adaptation for different property types or inspection requirements.
\end{itemize}

However, HRL implementation faces significant challenges. The credit assignment problem becomes substantially more complex when actions at multiple hierarchical levels contribute to eventual outcomes \citep{gao2023}. Defining appropriate interfaces between hierarchical levels requires careful design to balance abstraction with necessary information flow. Recent work on dynamic task allocation \citep{liu2025a} shows promise in automatically learning these hierarchical decompositions from data rather than requiring manual specification.

\subsection{The 3DGS Revolution in Active Reconstruction}

The adoption of 3D Gaussian Splatting has catalyzed a paradigm shift in active reconstruction, particularly for time and battery-constrained indoor inspection tasks. Unlike Neural Radiance Fields (NeRFs) that require computationally expensive volumetric rendering incompatible with real-time operation \citep{pan2024, yan2025}, 3DGS enables real-time reconstruction and rendering on drone-compatible hardware. This efficiency breakthrough has enabled entirely new approaches to intelligent view planning that were previously computationally infeasible.

Recent systems leverage the unique properties of 3DGS for sophisticated view selection strategies. ActiveGAMER directly optimizes viewpoint selection based on predicted improvements to rendered image quality, using the differentiable nature of Gaussian Splatting to estimate information gain \citep{chen2025e}. GauSS-MI formulates mutual information objectives directly in the Gaussian primitive domain, enabling efficient computation of expected information gain for candidate viewpoints \citep{xie2025}. AG-SLAM employs Fisher Information Matrix analysis of Gaussian parameters to balance exploration objectives with localization uncertainty minimization \citep{jiang2024a}.

A particularly innovative approach demonstrates that ranking potential views in the frequency domain can effectively estimate information gain without requiring ground truth data, achieving state-of-the-art results in view selection efficiency. For property inspection applications, this means drones can autonomously identify observation angles that will maximally improve understanding of room layouts and surface conditions, rather than following predetermined flight paths that may miss critical details or waste time on redundant observations.

The explicit nature of 3DGS representations enables novel uncertainty quantification approaches essential for inspection completeness verification. Unlike implicit neural representations, each Gaussian primitive can maintain associated confidence metrics reflecting observation quality and reconstruction fidelity. This enables systems to identify and prioritize revisiting high-uncertainty regions---crucial for documenting complex damage patterns that require multiple viewing angles for accurate assessment.

Practical integration of 3DGS into operational inspection systems addresses real-world deployment challenges beyond laboratory demonstrations. GS-Planner exemplifies this integration by implementing online quality evaluation of reconstructed 3DGS maps to guide exploration while maintaining safety constraints suitable for quadrotor navigation in cluttered environments \citep{jin2024}. ActiveGS advances this concept by maintaining explicit per-primitive confidence scores throughout the reconstruction process \citep{jin2025}, enabling targeted acquisition strategies that specifically address regions of high uncertainty.

These systems transform drones from passive scanners mechanically following predetermined routes into intelligent agents that actively seek maximally informative viewpoints based on evolving scene understanding. For property assessment, this translates directly into shorter flight times, more complete coverage of critical areas, and higher confidence in defect detection---addressing the fundamental economic and quality challenges of drone-based inspection services.

\subsection{Collaborative Swarms: Scaling Through Parallelism}

The emergence of collaborative drone swarms \citep{chung2018} marks a fundamental shift in autonomous inspection capabilities, demonstrating how parallel coordination can dramatically reduce task completion times. Recent investigations have shown that heterogeneous teams achieve significantly improved target visibility while operating in substantially compressed timeframes compared to single-drone deployments (\citep{nathan2023, freda2023, dhami2024, tang2025c}). These technological advances now enable zero-human-entry inspections in confined spaces while simultaneously reducing operational costs, fundamentally altering how infrastructure monitoring is conducted.

\subsubsection{The Mathematics of Swarm Efficiency}

Recent advances in collaborative simultaneous localization and mapping (C-SLAM) reveal that parallelism in robotic systems follows predictable efficiency patterns. A particularly striking development comes from ETH Zurich's ultra-lightweight system, which achieves sub-decimeter mapping accuracy using nano-UAVs despite minimal computational resources---operating with memory requirements several orders of magnitude below conventional systems \citep{zhou2022}. This miniaturization breakthrough opens new possibilities for swarm deployment in scenarios where traditional high-performance platforms would be impractical or economically unfeasible.

Complementing these hardware advances, decentralized frameworks like Swarm-SLAM demonstrate how communication overhead can be reduced through sparse representations while maintaining mapping accuracy \citep{lajoie2024}. Systems such as Kimera-Multi push the boundaries further by enabling real-time collaborative construction of metric-semantic 3D meshes across robot teams \citep{tianKimeraMultiRobustDistributed2021}. The RACER algorithm represents another significant advance, achieving fully decentralized exploration with asynchronous communication and demonstrating the first real-world deployment of multi-UAV collaborative exploration \citep{zhou2023}.

The efficiency gains from swarm deployment follow patterns of diminishing returns that reach optimal points at specific drone densities (\citep{vsrhelyi2018, zhou2022, ariasperez2025}). In confined space applications, even dual-drone systems can provide substantial improvements over single-drone operations \citep{nathan2023}. The coordination overhead inherent in swarm systems eventually begins to dominate the parallelization benefits, though the exact point varies with the swarm framework employed \citep{alqudsi2025}. However, larger swarms retain clear advantages for expansive outdoor inspections where communication latency has less impact on overall performance \citep{zhu2023}. Communication bandwidth consistently emerges as the primary scaling limitation, with WiFi mesh networks supporting substantially larger simultaneous drone operations compared to ultra-wideband systems \citep{alqudsi2025a}.

\subsubsection{Heterogeneous Architectures Unlock Specialized Efficiency}

The transition from homogeneous to heterogeneous swarms represents a fundamental evolution in inspection strategy (\citep{tranzatto2022, zafar2025, liu2025a}). Leader-follower architectures, exemplified by the SwarmGear system, illustrate how specialized roles can substantially enhance overall performance. In this configuration, leader drones equipped with compliant robotic legs enable seamless aerial-to-ground transitions while maintaining precise formation with followers through virtual impedance links \citep{darush2023}.

\begin{figure*}
    \centering
    \includegraphics[width=1\textwidth]{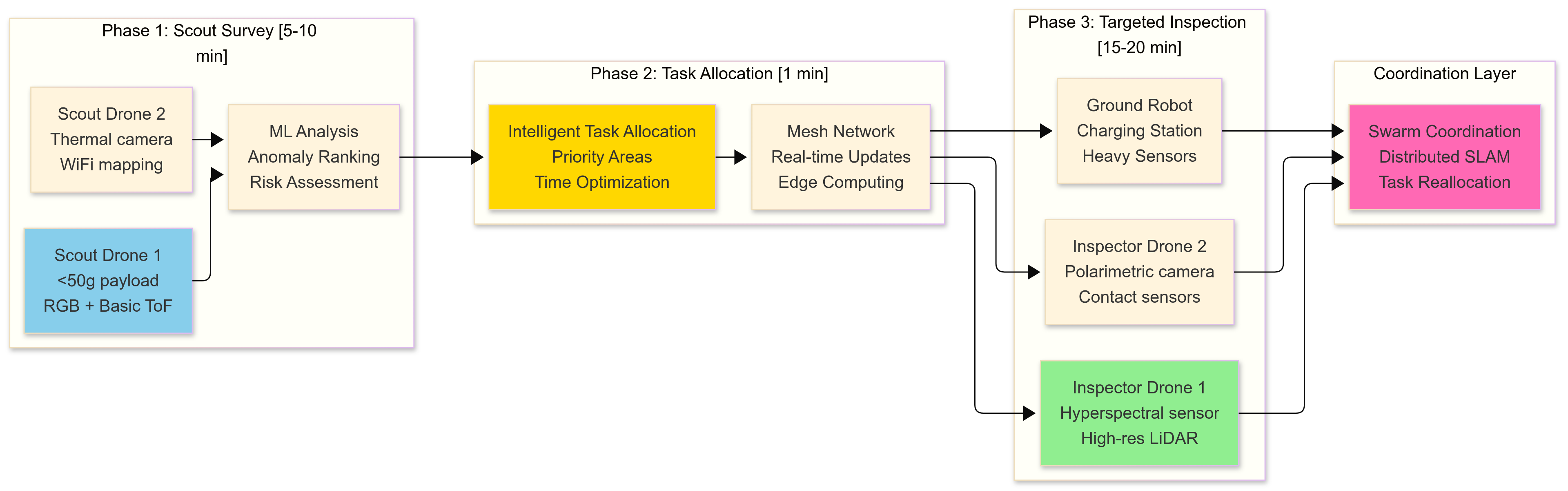}
    \caption{Example of a heterogeneous swarm with different roles for different types of drones. Phase 1 (5-10min): Scout drones (<50g) with RGB/thermal/ToF rapidly survey and ML-rank anomalies. Phase 2 (1min): Task allocation optimizes targets across drone capabilities using mesh networking. Phase 3 (15-20min): Inspector drones deploy specialized sensors (hyperspectral, polarimetric, LiDAR) while ground robots access confined spaces and provide charging. Distributed SLAM maintains consistent maps despite individual limitations.}
    \label{fig:scout-inspector}
\end{figure*}

Scout-inspector paradigms provide another compelling example of functional specialization benefits (\autoref{fig:scout-inspector}). Lightweight detection UAVs rapidly survey areas using minimal sensor payloads (\citep{zhou2021, bartolomei2023, zhou2023, zheng2025a, brugali2025}), while more capable inspection drones equipped with high-resolution cameras and specialized sensors conduct detailed analysis of identified areas (\citep{francos2023, jacobsen2023, cao2025}). This division of labor eliminates redundant capabilities and optimizes energy consumption---scout drones can operate for extended periods on minimal battery capacity while inspectors deploy power-intensive equipment only when and where needed. The concept extends to distributed computing architectures, where certain drones in the swarm handle computationally intensive tasks on behalf of data acquisition units \citep{ramshanker2024}, demonstrating the versatility of heterogeneous design principles.

The integration of unmanned ground vehicles (UGVs) with aerial swarms creates inspection capabilities that neither platform could achieve independently. In building inspection scenarios, UAV-UGV collaboration has demonstrated measurable improvements in mission efficiency, with significant reductions in average completion times \citep{munasinghe2024}. The complementary nature of these platforms---UAVs providing rapid overview and exterior access while UGVs handle detailed ground-level inspection and serve as mobile charging stations---enables continuous operations that would be impossible with single-platform systems \citep{munasinghe2024}.

Recent advances in language-guided heterogeneous teams further expand operational possibilities. Natural language mission specification allows operators to communicate intent rather than low-level control commands to UAV-UGV teams \citep{cladera2025}. These systems employ Large Language Model (LLM)-enabled planners that reason over semantic-metric maps built online and opportunistically shared between aerial and ground robots, successfully demonstrating kilometer-scale navigation in both urban and rural environments \citep{cladera2025}. The SPINE framework exemplifies practical deployment of LLM-enabled autonomy in field robotics, addressing the challenges of large-scale unstructured environments through active exploration and navigation of obstacle-cluttered terrain \citep{ravichandran2025a}. Notably, this framework's model-agnostic approach enables model distillation for size, weight, and power (SWaP) limited platforms, achieving the first language-driven UAV planner using on-device language models \citep{ravichandran2025a}. These developments effectively bridge the gap between high-level mission specification and low-level robot coordination, enabling heterogeneous swarms to adaptively respond to changing mission requirements in real-time.

\subsubsection{From Laboratory to Infrastructure: Quantified Real-World Gains}

The transition from research prototypes to operational systems \citep{aloui2024} yields compelling performance metrics across multiple application domains. Bridge inspection operations that previously required full-day manual efforts with extensive scaffolding and multiple workers can now be completed in hours by single operators using automated swarm systems \citep{panigati2025}. Power line inspection capabilities demonstrate order-of-magnitude improvements compared to traditional ground crews, with swarm deployments enabling simultaneous multi-section coverage that further multiplies efficiency gains \citep{jacobsen2023}.

The economic impact proves equally significant. Traditional inspection methods requiring specialized equipment and extensive personnel are being replaced by drone swarms that eliminate equipment rental costs and reduce staffing requirements by more than half \citep{jacobsen2023}. Organizations implementing drone inspection programs report substantially decreased insurance costs, reflecting the dramatic reduction in worker exposure to hazardous environments \citep{aws2025}.

Quality improvements accompany these efficiency gains. Modern swarms achieve exceptional accuracy in defect detection, with specialized sensors capable of identifying submillimeter-scale features. The comprehensive 3D models generated through collaborative mapping provide persistent digital records that enable trend analysis and predictive maintenance---capabilities that were entirely absent from traditional inspection methods.

\subsubsection{Biomimetic Miniaturization Conquers Confined Spaces}

The implementation of zero-human-entry policies for confined space inspection has catalyzed remarkable innovations in miniaturized robotics. Major industrial stakeholders now mandate complete elimination of human entry into hazardous environments, driving the development of robots capable of navigating extremely confined spaces while maintaining full inspection capabilities. These advanced systems integrate multiple locomotion modes---using aerial navigation to reach access points followed by surface adhesion through bio-inspired mechanisms \citep{voliro2025}.

China's Jiu Tian aerial mothership, scheduled for 2025 deployment, represents an extreme manifestation of the mothership concept. This 10-ton platform can deploy 100 smaller drones simultaneously across a 7,000-kilometer operational range, functioning as an airborne command center that coordinates swarm operations while providing power and communication relay capabilities \citep{magazine2025}. For more routine commercial applications, lighter-than-air blimp carriers offer extended loiter capabilities with minimal energy consumption, creating persistent aerial platforms suitable for continuous monitoring operations \citep{bhat2024a, cooney2025}. The mothership concept scales down as well, with smaller traditional drones serving as carriers \citep{sorbelli2024}, as demonstrated in last-mile delivery applications by companies like Amazon and Manna Aero.

The convergence of miniaturization and swarm coordination enables entirely new inspection paradigms. Quadcopter drones dropped from blimp motherships can carry even smaller nanobots \citep{liu2024d} capable of navigating ventilation systems while buildings remain fully occupied. Wall-climbing robots now achieve practical climbing speeds on vertical surfaces, while biomimetic soft actuators can conform to complex geometries that would be inaccessible to rigid robots \citep{ma2024, fitzgerald2025}.

\subsubsection{Business Model Innovation: From Ownership to Outcomes}

The Robotics-as-a-Service (RaaS) model fundamentally restructures the economics of robotic inspection. Rather than requiring substantial capital investment in hardware and software, organizations can now access advanced swarm capabilities through subscription-based models \citep{chen2010, jacquillat2024}. This transition from ownership to outcomes accelerates technology adoption while effectively transferring technology risk from end users to service providers.

Fleet management platforms provide the essential software infrastructure for scalable operations. These cloud-based systems handle mission planning, real-time coordination, regulatory compliance tracking, and predictive maintenance scheduling \citep{singhal2017}. By reducing operational complexity while providing enterprise-grade reliability, they enable smaller inspection companies to compete effectively with established players while helping large organizations standardize operations across geographically distributed portfolios.

The economics prove compelling across all deployment scales. Smaller operators can achieve rapid return on investment through reduced labor costs and increased inspection frequency, while large-scale deployments benefit from volume efficiencies and advanced coordination capabilities. These economies of scale often result in substantially lower per-inspection costs for larger fleets, creating natural market consolidation dynamics that favor sophisticated operators.

As swarm robotics transitions from research demonstrations to operational necessity, the convergence of technical capability, economic viability, and regulatory acceptance creates unprecedented opportunities for infrastructure monitoring \citep{fernandezcortizas2024, widhalm2025}. Organizations embracing collaborative swarm architectures position themselves to capitalize on rapidly expanding markets. The fundamental question has evolved from whether robotic swarms will transform infrastructure inspection to how quickly traditional methods will become obsolete. The mathematical principles of parallelism, combined with specialization through heterogeneous architectures and accessible business models, ensure that collaborative swarms represent not merely incremental improvement but a fundamental reimagining of how we monitor and maintain critical infrastructure.

\section{Integration with the Built Environment}

\subsection{Building Information Modeling: From Static Design to Dynamic Reality}

The convergence of physics-aware 3D Gaussian Splatting with Building Information Modeling represents a pivotal development in how autonomous drones understand and interact with constructed spaces \citep{ruter3DGaussianSplatting2024}. Traditional BIM systems excel at representing design intent---the theoretical state of a building as conceived by architects and engineers. However, the physical reality of buildings diverges from these idealized models through construction variations, material degradation, and ongoing modifications. Digital Twin technology extends BIM's capabilities by establishing continuous bi-directional communication between digital models and physical assets, enabling real-time monitoring throughout a building's operational life \citep{tagliabue2024}.

The integration of spectropolarimetric 3DGS pushes this concept further by capturing not just geometric accuracy but fundamental material properties. Where conventional scanning methods record surface appearance, spectropolarimetric approaches reveal what materials actually \emph{are}---their chemical composition, moisture content, and structural integrity (\citep{kim2023, li2024h, thirgood2024}). This creates what might be termed ``verifiable physical digital twins''---models that encode not just shape and color, but the underlying physical truth of building materials.

\subsubsection{Graph-Based Architectures and Semantic Evolution}

The upcoming IFC 5 standard marks a significant architectural shift in how building data is structured and accessed. Drawing inspiration from the gaming industry's approach to complex scene management, IFC 5 abandons the monolithic file structures that have constrained previous versions in favor of granular, componentized data that supports distributed collaboration and real-time updates (\citep{github2025, buildingsmart2025, biblus2025}). This architectural evolution directly addresses a critical limitation: traditional BIM schemas struggle to incorporate continuous sensor streams \citep{ma2025, wang2025f}, making real-time integration of physics-aware 3DGS data practically impossible.

Graph database implementations offer elegant solutions to these integration challenges. Recent research demonstrates how graph structures can encode complex spatial relationships that traditional relational databases struggle to represent \citep{zhu2023a, iranmanesh2025}. These systems excel at extracting implicit relationships---identifying which rooms connect through doors, which walls share structural loads, or which spaces depend on specific HVAC zones \citep{zhu2025}.

Perhaps most intriguingly, natural language query capabilities are emerging that allow users to ask questions like ``show me all concrete walls with moisture levels above 15\%'' or ``identify areas where actual materials differ from specifications'' \citep{blog2025}. These queries combine semantic understanding of building components with real-time sensor data, enabling investigations that would require hours of manual analysis using traditional tools \citep{laina2025}. Practical implementations demonstrate how nodes can represent discrete building elements while edges capture both physical connections and logical relationships like ``supplies power to'' or ``thermally influences'' \citep{tong2025}.

The transition to semantic representations faces notable obstacles. The ifcOWL experiment---an attempt to express IFC schemas as web ontologies---revealed fundamental incompatibilities between EXPRESS-based schemas and semantic web technologies. The resulting ontologies required numerous exceptions and workarounds that complicated implementation and limited practical adoption \citep{levy2014, technical2025}. However, federated knowledge graph approaches show promise by allowing different domains to maintain their native representations while establishing semantic bridges for data exchange.

\subsubsection{Robotic Scan-to-BIM Transformation}

The integration of autonomous robotics with BIM workflows represents a qualitative leap beyond traditional scanning methodologies (\citep{he2024a, chen2025f, liu2025b}). Where conventional approaches rely on human operators positioning scanners at predetermined locations, robotic systems make intelligent decisions about viewpoint selection based on accumulated knowledge and identified uncertainties. Research demonstrates that BIM-enriched navigation schemas enable robots to achieve coverage rates exceeding 95\% of visible surfaces while maintaining positional accuracy within centimeters \citep{zhai2024a}.

Construction sites present particularly challenging environments for autonomous scanning, with temporary structures, moving equipment, and evolving geometries. Recent work on automated scaffold scanning shows how robots can navigate these complex environments, achieving more comprehensive coverage than human operators who may miss occluded areas or avoid potentially hazardous locations \citep{chung2025}.

The emergence of AI-powered tools that generate BIM models directly from Gaussian Splatting data represents a convergence of photogrammetric and semantic understanding \citep{xgrids2025}. These systems don't merely reconstruct geometry---they identify building elements, classify materials, and establish relationships that populate BIM databases. The adoption of JSON-based serialization in IFC 5 facilitates this integration by providing flexible schemas that can accommodate the rich spectropolarimetric data streams generated by advanced sensors \citep{biblus2025}.

This transforms Scan-to-BIM from a primarily geometric exercise into comprehensive material verification \citep{mahmoud2024}. Imagine a drone scanning a newly constructed wall: beyond capturing its dimensions and position, spectropolarimetric sensors detect the actual concrete composition, identify areas where mixture ratios deviate from specifications, and flag locations where inadequate curing has compromised strength \citep{liuSemanticGaussianSplattingenhanced2025}. This information feeds directly into BIM systems, creating as-built models that reflect not just what was designed, but what was actually constructed.

\subsubsection{Context-Aware Infrastructure Intelligence}

The fusion of BIM data with real-time sensor streams enables a new paradigm of context-aware inspection. Traditional inspection routes follow predetermined patterns, often redundantly scanning areas in good condition while potentially missing emerging problems. BIM-integrated systems incorporate both static design knowledge and dynamically accumulated inspection results to generate adaptive scanning strategies that focus attention where it's most needed \citep{liu2025c}.

Semantic reasoning tools can now automatically infer relationships that were never explicitly modeled. For instance, understanding that a water leak in one location likely affects materials in adjacent spaces below, or recognizing that structural cracks in load-bearing elements require immediate investigation of connected components \citep{lilis2025}. These inference capabilities break down the traditional silos between architectural, structural, and mechanical models.

Heterogeneous robotic teams exemplify how BIM knowledge enhances collaborative inspection. Quadruped robots like Boston Dynamics' Spot serve as mobile bases for deploying specialized drones, combining ground-level stability with aerial access \citep{asadi2020, darush2023}. The quadruped navigates using BIM-derived maps, identifying optimal deployment locations for drones to inspect facades, ceilings, or confined spaces \citep{munasinghe2024}.

This collaborative approach leverages each platform's strengths: ground robots provide extended operation time and carry heavy sensors, while drones offer rapid deployment and access to otherwise unreachable areas \citep{yeoh2023}. Commercial systems like Ghost Robotics' Vision 60 demonstrate practical implementations with modular payload systems supporting various sensor configurations \citep{robotics2024}. Studies show such heterogeneous teams can reduce inspection time by up to 60\% while achieving more thorough coverage than single-platform approaches \citep{munasinghe2024}.

The integration extends beyond inspection to active building management. Machine learning algorithms analyze patterns in BIM data, sensor readings, and environmental conditions to optimize HVAC operations, reducing energy consumption while maintaining occupant comfort \citep{iqbalDigitalTwinEnabledBuilding2025}. Advanced lighting control systems coordinate natural daylight with artificial illumination, using BIM geometry to predict shadow patterns and optimize fixture activation \citep{esser2024}.

\subsubsection{Physical Digital Twin Implementation}

Real-world deployments reveal both the promise and challenges of BIM-integrated Digital Twins. A comprehensive case study of a commercial building achieved remarkable temporal synchronization by embedding diverse sensors throughout the structure---vibration monitors in structural elements, environmental sensors in occupied spaces, and water detection systems in vulnerable areas \citep{yang2025a}. The key innovation wasn't just sensor deployment but the creation of a unified data architecture that maintains coherent relationships between physical measurements and BIM elements.

Performance optimization remains crucial for real-time applications. Traditional BIM databases struggle with the velocity and volume of continuous sensor streams. Recent advances using in-memory graph databases and optimized query structures demonstrate update rates exceeding 1000 Hz---fast enough to capture structural vibrations and enable real-time health monitoring \citep{eneyew2022, assaad2024}.

Hybrid architectures offer practical solutions by maintaining complete geometric and semantic data in relational databases while using lightweight graph representations for rapid traversal and relationship queries \citep{yue2025}. This approach reduces query times from minutes to milliseconds---essential for autonomous drones making real-time navigation decisions based on structural knowledge.

\subsubsection{USD and IFC5: Convergent Technologies}

The influence of Universal Scene Description (USD) on IFC 5's development reflects broader trends in 3D data representation. Originally developed by Pixar for managing complex animated scenes, USD's layering system and non-destructive editing paradigm offer elegant solutions to long-standing BIM challenges \citep{buildingsmart2024, bimgym2025}).

Zaha Hadid Architects' implementation of USD-based workflows demonstrates practical benefits: projects with millions of polygons that previously required overnight processing now update in real-time, enabling fluid design iteration and immediate visualization of changes \citep{nvidia2025a}. The key insight is treating building information as composed layers---structural systems, mechanical equipment, sensor data---that can be independently modified without disrupting other domains.

This layered approach elegantly resolves the tension between prescriptive BIM models and observational reality. The base IFC layer maintains design intent and regulatory compliance information. Dynamic layers overlay real-time sensor data, inspection results, and performance metrics. Graph databases serve as the semantic backbone, maintaining relationships across layers while enabling complex queries that span multiple domains. Though computational demands remain challenging for real-time indoor navigation, successful implementations demonstrate feasibility with appropriate optimization.

\subsubsection{Intelligent Generation and Autonomous Systems}

The evolution from passive documentation to active generation marks a fundamental shift in BIM's role. Graph neural networks trained on thousands of building designs can now generate floor plans that not only satisfy spatial requirements but automatically create compliant BIM models with properly classified elements and code-conforming relationships \citep{liu2024e}. This isn't simple automation---it represents machine understanding of architectural principles and building codes.

Process mining techniques applied to construction data reveal hidden patterns in project execution. By analyzing temporal sequences in BIM modifications, sensor data, and inspection records, algorithms can predict likely completion dates, identify recurring bottlenecks, and suggest optimizations that have improved project efficiency by up to 20\% in pilot studies \citep{panBIMdataMiningIntegrated2021}. Risk management systems leverage these patterns to flag potential issues before they manifest, transitioning from reactive problem-solving to proactive prevention \citep{huang2024a}.

For property assessment applications, this integration enables a fundamental shift from subjective evaluation to objective measurement. Rather than an inspector noting ``basement exhibits evidence of dampness,'' integrated systems provide quantitative assessments: ``Basement drywall surfaces show 15\% moisture content increase indicating water damage initiated approximately 6 months ago based on spectral degradation patterns. Laminate delamination detected in 30\% of door surfaces. Material composition varies from specifications by 12\%, suggesting non-original replacements.'' This precision not only supports more accurate valuations but provides defensible documentation for insurance claims and legal proceedings.

The convergence of BIM with physics-aware sensing represents more than technical integration---it fundamentally reimagines how we understand and interact with built environments. As autonomous drones equipped with spectropolarimetric sensors become standard tools, the artificial distinction between digital models and physical buildings dissolves. In its place emerges a unified representation that captures not just what was designed or what appears to exist, but the verifiable physical truth of our built environment.


\subsection{From Subjective Assessment to Objective Measurement}

The transformation of property assessment from art to science addresses a fundamental measurement problem that costs billions annually. While financial markets have embraced quantitative analysis and algorithmic trading, real estate---humanity's largest asset class---remains largely dependent on subjective visual inspection. Physics-aware sensing technologies promise to change this, offering precise, defensible measurements where previously only opinions existed.

\subsubsection{Uncovering Hidden Water Damage: The Invisible Destroyer}

Water damage represents one of the most costly and contentious issues in property assessment, generating approximately \$13 billion in annual insurance claims \citep{affairs2025}. The insidious nature of water damage---often progressing invisibly within walls and under floors---means traditional detection methods identify problems only after significant structural damage has occurred. By this point, simple repairs have escalated into major renovations.

Hyperspectral imaging fundamentally changes this dynamic by detecting water at the molecular level. Water molecules exhibit characteristic absorption patterns in specific near-infrared wavelengths, creating spectral signatures as distinctive as fingerprints \citep{divyaSoilWaterContent2019}. Advanced sensors can identify moisture content changes as small as 2-3\%, well below the threshold for visible damage or even many moisture meters. This sensitivity enables detection of slow leaks within days rather than months, potentially saving thousands in repair costs.

Beyond early detection, spectral analysis introduces forensic capabilities that resolve longstanding disputes in property transactions and insurance claims. Different types of water damage create distinct spectral evolution patterns---a sudden pipe burst saturates materials differently than gradual seepage, leaving characteristic signatures in how moisture distributes through materials over time \citep{specim2021}. Research demonstrates that hyperspectral time-series analysis can effectively ``rewind'' moisture events, establishing timelines that differentiate pre-existing conditions from recent damage \citep{moghadam2021}. This capability transforms contentious ``he-said-she-said'' insurance disputes into objective determinations based on physical evidence.

In today's fast‑moving, post‑pandemic real estate market, buyers frequently waive the home‑inspection contingency, mistakenly believing that a lender's appraisal provides the same forensic scrutiny as a professional inspection. Yet USPAP limits appraisers to conditions that are ``readily observable,'' and the typical field kit doesn't include a moisture meter, FLIR imager, or borescope---so slow‑moving leaks hidden behind drywall often go unreported. When that latent damage finally surfaces, repair bills can erode reserves and push loans into distress; a recent study of FreddieMac mortgages shows that hurricane‑driven water damage raises the 180‑day default probability by about\,0.5\,percentage points---roughly a 70\,\% jump versus unaffected loans \citep{gete2024}. Integrating drone‑borne hyperspectral moisture mapping into the appraisal workflow would restore the protection buyers surrender when they waive inspections, forcing remediation or repricing before funding and materially lowering future buy‑back and default risk across the mortgage pool.

\subsubsection{Material Fraud: The Growing Challenge of Sophisticated Forgeries}

Modern manufacturing has produced remarkably convincing imitations of premium materials. Large‑format, gauged porcelain slabs printed with photorealistic veining routinely masquerade as Calacatta or Carrara marble; their visible‑light reflectance can differ from natural carbonate stone by less than\,0.5\%, leaving even veteran stone fabricators reliant on acid‑etch or spectroscopic tests for verification (\citep{asdrubaliThermalOpticalCharacterization2015} Because authentic Italian marble commands \$40--\$180 per square foot more than porcelain, such misidentification can distort collateral value by tens of thousands of dollars in a high-end renovation.

Spectral and polarimetric sensors pierce through surface appearances to reveal fundamental material properties. Wood possesses unique spectral signatures from lignin and cellulose that no synthetic material can replicate \citep{wang2024g}. Polarimetric imaging adds another dimension by measuring how materials scatter light---the ordered crystalline structure of natural stone creates polarization patterns distinctly different from manufactured alternatives \citep{han2023}.

These detection capabilities extend beyond countertops to flooring, fixtures, and even structural materials where substitution of specified products could compromise safety or longevity. The technology creates an authentication system that cannot be defeated by surface treatments or clever manufacturing---physics doesn't lie.

\subsubsection{Predictive Degradation: Moving from Reactive to Proactive}

Traditional property maintenance operates on failure-driven cycles: components work until they break, triggering expensive emergency repairs and potential consequential damage. Physics-aware sensing enables a predictive paradigm by detecting material degradation at the molecular level, long before visible symptoms appear.

Polymeric materials in roofing, siding, and seals undergo characteristic changes as they age---plasticizers migrate, molecular chains break, and UV damage accumulates. Hyperspectral sensors detect these chemical changes through subtle shifts in absorption spectra \citep{paoliniEffectAgeingSolar2014}. Concrete carbonation, a slow process that eventually compromises structural integrity, creates detectable changes in surface chemistry years before spalling or cracking occurs. Polarimetric techniques can even measure microscopic surface roughness changes that indicate early stages of material breakdown \citep{kupinski2022}.

This predictive capability enables entirely new maintenance strategies. Rather than replacing roofs on fixed schedules or after leaks appear, building owners can monitor actual degradation rates and optimize replacement timing. Insurance companies can price policies based on measured material conditions rather than crude age-based models. Warranty claims can be validated through objective degradation measurements rather than subjective assessments.

\subsubsection{Adoption Barriers: The Inertia of Established Practice}

Despite compelling technical capabilities, physics-aware sensing faces significant adoption challenges rooted in institutional inertia and regulatory frameworks designed around traditional methods. The appraisal industry, governed by the Uniform Standards of Professional Appraisal Practice (USPAP), emphasizes consistency and comparability in valuation methods \citep{taf2024}. While USPAP requires appraisers to consider ``relevant evidence,'' the interpretation of what constitutes acceptable evidence has historically favored established data types.

Current reporting standards like the Uniform Appraisal Dataset (UAD) 3.6 lack fields for spectral signatures or material composition data \citep{mae2025}. Appraisers trained to assign condition ratings like ``C3'' (average condition) have no standardized way to incorporate quantitative moisture readings or material verification data. This creates a chicken-and-egg problem: technology providers hesitate to develop solutions without clear market demand, while institutions resist adoption without proven integration paths.

Legal systems present similar challenges. Courts rely heavily on precedent and established evidentiary standards. Introducing spectral analysis as evidence requires expert testimony to establish reliability, relevance, and interpretation methods. Early adopters must invest in educating judges, juries, and opposing counsel about the scientific basis and practical implications of physics-based measurements.

\subsubsection{Catalyzing Adoption Through Demonstrated Value}

Successful technology adoption requires clear value propositions for each stakeholder in the property ecosystem. For appraisers, physics-aware sensing doesn't replace professional judgment but provides objective support for subjective assessments. An appraiser noting ``C3'' condition can now append quantitative evidence: ``Rating supported by spectropolarimetric scan showing moisture levels below 8\% throughout, original material specifications confirmed, surface wear consistent with 5-7 years normal use.'' This transforms appraisals from opinions into scientifically-backed assessments.

Insurance companies face immediate, quantifiable benefits. With water damage claims averaging \$13,954 in severity between 2018-2022 \citep{house2025}, even modest improvements in claim accuracy yield substantial savings. Pre-loss documentation using spectral scans eliminates disputes about prior conditions. Claims adjusters can differentiate sudden covered events from gradual excluded deterioration using spectral timeline analysis. Subrogation opportunities improve when physical evidence establishes causation and timing.

Commercial implementations demonstrate practical feasibility. The Specim AFX drone platform already performs hyperspectral surveys for precision agriculture and environmental monitoring (\citep{sassi2023, wolff2024, kourounioti2025}). Adapting these systems for indoor property inspection requires engineering solutions for confined spaces and artificial lighting, but no fundamental technological breakthroughs. Early adopters in commercial property management report 15-20\% reductions in maintenance costs through predictive intervention enabled by continuous monitoring.

Mortgage lenders and secondary market participants have compelling incentives for adoption. The 2008 financial crisis highlighted risks from inadequate property documentation, leading to billions in buyback demands. Physics-based verification of property condition and materials provides defensible documentation that reduces origination risk and supports accurate securitization ratings. Detecting undisclosed damage or material substitutions before loan closing protects against value impairment that could trigger future losses.

\subsection{The Platform Model: Network Effects and Data Arbitrage}

The integration of physics-aware sensing with autonomous inspection platforms creates conditions for powerful network effects that characterize successful technology platforms \citep{margolis2023}. Unlike traditional inspection services that deliver discrete reports, continuous sensing generates accumulating data assets that appreciate in value over time.

\subsubsection{Algorithmic Learning and Competitive Moats}

Each property scan contributes to expanding databases of material signatures, degradation patterns, and failure modes. Machine learning algorithms trained on thousands of moisture damage cases can predict failure probability with increasing accuracy as data accumulates \citep{haftorHowMachineLearning2021}. A platform that has analyzed 100,000 kitchen cabinets possesses predictive capabilities no new entrant can match without similar data access. This creates powerful first-mover advantages and barriers to entry.

The network effects extend beyond simple data accumulation. As platforms identify correlations between materials, environmental conditions, and degradation rates, they can offer increasingly sophisticated services \citep{vomberg2023, climent2024}. Predictive models might reveal that certain vinyl flooring formulations degrade rapidly in high-humidity coastal environments, or that specific concrete mixtures show accelerated carbonation in urban areas with elevated CO2 levels. These insights enable targeted inspections, optimized maintenance schedules, and risk-adjusted pricing that create value for all ecosystem participants.

\subsubsection{Workflow Integration and Switching Costs}

As appraisers, inspectors, and property managers integrate physics-based data into their workflows, switching costs naturally emerge \citep{toprakli2025}. Software platforms like Xactimate that currently rely on visual assessment and manual measurement could incorporate spectral verification modules. Appraisal reports enhanced with objective sensor data become the new standard, making traditional assessments appear incomplete or less credible.

The integration process creates technical and behavioral lock-in effects \citep{hofmann2025, wang2025h}. Professionals invest time learning to interpret spectral data and incorporate findings into reports. Organizations develop processes around automated alerts for moisture detection or material degradation. Historical data becomes valuable for trend analysis and comparison. These accumulated investments in platform-specific knowledge and processes create natural retention even as competitors emerge.

\subsubsection{Regulatory Alignment and Market Creation}

Rather than fighting existing standards, successful platforms will align with and enhance regulatory compliance \citep{dabestani2025}. USPAP's requirement for ``relevant evidence and logic'' creates natural demand for objective measurements that support subjective assessments \citep{foundation2024}. Forward-thinking platforms will work with standard-setting bodies to establish acceptable methods for incorporating spectral data into traditional reporting frameworks.

This regulatory alignment transforms potential resistance into market pull. As courts accept spectral evidence in property disputes, parties without such documentation face disadvantage. Insurance regulators recognizing the fraud-prevention benefits might incentivize or mandate physics-based verification for certain coverage types. Government agencies seeking to improve building safety could require spectral verification of material compliance in critical applications.

\subsubsection{Data Arbitrage and Information Asymmetry}

The ultimate value creation mechanism lies in systematic identification of mispriced assets through superior information. A platform with comprehensive spectral databases can identify properties where hidden damage or material substitution creates valuation errors. This information asymmetry enables various monetization strategies: direct property investment, information services to institutional investors, or risk assessment for lenders and insurers.

Consider a platform that scans thousands of properties monthly. Statistical analysis might reveal that 5\% show significant undisclosed water damage, 3\% have non-conforming materials affecting value, and 2\% exhibit accelerated degradation requiring near-term capital expenditure. This knowledge enables profitable strategies: acquiring undervalued properties with correctible defects, shorting mortgage securities with concentrated exposure to compromised properties, or offering targeted insurance products with appropriate risk pricing.

\subsection{Implications for Autonomous Systems Research}

The property assessment domain offers unique challenges that push the boundaries of current robotics research while providing clear commercialization paths that address the traditional ``valley of death'' between academic research and market deployment \citep{beard2009, cho2025}.

\subsubsection{Unstructured Environment Navigation}

Unlike industrial robots operating in controlled environments, property inspection demands navigation through diverse, cluttered, and unpredictable spaces \citep{leeSurveyRoboticsTechnologies2023, macaulay2022}. Each home presents unique layouts, furniture arrangements, and obstacles. Autonomous drones must reason about traversability, maintain safe distances from valuable objects, and adapt to unexpected conditions like closed doors or moved furniture \citep{chen2024g, gao2019}.

This challenge drives advances in semantic understanding and adaptive planning. Rather than following predetermined paths, inspection drones must understand space at a conceptual level---recognizing that kitchens require detailed moisture scanning, that bathrooms present confined navigation challenges, or that historic properties demand extra care around irreplaceable features. This pushes research toward embodied AI systems that combine perception, reasoning, and action in sophisticated ways \citep{liuAligningCyberSpace2024}.

\subsubsection{Multi-Modal Sensor Fusion Under Constraints}

The integration of visual, thermal, hyperspectral, and polarimetric sensors on weight-limited platforms presents fundamental challenges in computational imaging and signal processing. Each modality generates massive data streams that must be processed in real-time for navigation and inspection tasks. Traditional approaches to multi-modal fusion quickly overwhelm the computational resources available on sub-500g platforms.

This constraint drives innovation in edge AI and efficient processing architectures. Researchers must develop algorithms that extract maximum information from minimal computation, using techniques like attention mechanisms to focus processing on informative regions, or hierarchical representations that enable coarse-to-fine analysis. The solutions developed for property inspection have broader applications in disaster response, environmental monitoring, and defense applications \citep{la2017, wu2025b}.

\subsubsection{Human-Robot Interaction in Sensitive Environments}

Property inspection occurs in personally significant spaces where trust and transparency become paramount. Homeowners allowing autonomous drones into their homes need confidence in privacy protection, predictable behavior, and clear communication about findings. This creates research opportunities in explainable AI, natural interaction paradigms, and trust-building through transparent operation.

The development of interfaces that communicate complex spectral findings to non-technical users pushes visualization and explanation research forward. How does a system explain that molecular-level moisture detection indicates probable pipe deterioration without causing unnecessary alarm? These challenges in human-centered AI design have implications across healthcare, education, and other domains where complex technical systems must interface with general users \citep{imranAIRoboticsEmbodied2025}.

\subsubsection{Bridging Perception and Physical Understanding}

The evolution from capturing appearance to measuring physical properties represents a fundamental advance in robotic perception \citep{capgemini2025}. Current computer vision excels at geometric reconstruction and object recognition but struggles with material understanding and physical property estimation. Physics-aware sensing pushes toward robots that understand not just what they see, but what objects are made of, how they degrade, and what risks they present.

This deeper understanding enables new capabilities: robots that predict which structures need reinforcement before failure, identify materials requiring special handling, or detect invisible hazards like chemical contamination. The fusion of geometric and material understanding creates opportunities for robots that interact more intelligently with their environment---a key step toward truly autonomous systems.

\section{Conclusion}

The convergence of autonomous indoor drones with physics-aware sensing technologies represents a watershed moment in how we understand, value, and maintain the built environment. Throughout this manuscript, we have explored the technical foundations that make this transformation possible---from the engineering constraints of sub-500g platforms to the sophisticated algorithms that transform spectral signatures into actionable insights about material degradation and structural integrity.

This technological evolution addresses a fundamental measurement crisis that has persisted since humans first began trading property: the gap between what we can see and what we need to know. While visual inspection reveals surface conditions, the technologies described here penetrate deeper, revealing moisture infiltration before stains appear, detecting material substitutions that fool the eye, and predicting failures while prevention remains economical. Hyperspectral imaging, polarimetric sensing, and their spectropolarimetric fusion don't merely enhance existing inspection methods---they redefine what inspection means.

The path from laboratory demonstration to widespread deployment faces genuine challenges. Sensor miniaturization must continue advancing to meet payload constraints. Processing algorithms need optimization for edge deployment. Regulatory frameworks require updating to accommodate new data types. Professional workflows must evolve to incorporate quantitative measurements. Yet none of these challenges appear insurmountable. Computational metasurfaces promise dramatic sensor size reductions. Neuromorphic processors enable efficient on-device AI. Early adopters in insurance and commercial property management demonstrate viable business models. The trajectory is clear, even if the timeline remains uncertain.

The implications extend far beyond technical achievements. In a world where property represents the primary store of wealth for most families, subjective assessment perpetuates inequalities and enables fraud. Physics-aware sensing promises greater objectivity---not perfect fairness, but meaningful progress toward decisions based on measurable reality rather than opinion or bias. When spectral signatures reveal water damage regardless of fresh paint, or polarimetric imaging confirms genuine hardwood despite convincing vinyl imitations, technology serves equity.

The transformation from hardware-defined to software-defined capabilities accelerates innovation cycles. Future inspection drones will gain new capabilities through algorithm updates rather than hardware replacement. Metasurfaces designed by AI will adapt their optical properties in real-time. Learned exploration policies will improve through experience across millions of inspections. Collaborative swarms will share insights and coordinate coverage. This shift from atoms to bits as the primary innovation substrate fundamentally changes development economics and competitive dynamics.

For researchers, this domain offers rich challenges at the intersection of perception, reasoning, and action. Unlike controlled laboratory environments, real properties present cluttered spaces, uncertain lighting, and skeptical occupants. Success requires advances in semantic understanding, efficient multi-modal fusion, explainable AI, and human-robot interaction. The clear commercial value provides motivation and funding often lacking in purely academic pursuits.

For industry practitioners, the message is clear: transformation is coming. Organizations that begin integrating physics-aware sensing into their workflows will develop competitive advantages through superior information and operational efficiency. Those that resist risk obsolescence as objective measurement becomes the expected standard. The question is not whether to adopt these technologies, but how quickly and strategically to do so.

For policymakers and standard-setting bodies, proactive engagement can shape this transformation to maximize societal benefit. Updated appraisal standards that incorporate spectral data would accelerate adoption. Insurance regulations recognizing physics-based documentation could reduce fraud and improve claim resolution. Building codes requiring material verification in critical applications would enhance safety. Thoughtful policy can channel technological capability toward public good.

As we stand at this inflection point, the vision is compelling: autonomous drones that see beyond human perception, revealing the true physical state of our built environment with unprecedented clarity. These systems will detect problems before they become catastrophes, verify quality without destructive testing, and provide objective evidence where previously only opinions existed. The transformation from photons to physics---from capturing light to understanding matter---represents more than technological progress. It embodies humanity's advancing ability to perceive, understand, and nurture the structures that shelter our lives and endeavors.

The foundations are laid. The technologies exist. The need is clear. What remains is execution---the dedicated effort to transform vision into reality through continued research, development, and deployment. As this transformation unfolds, we move toward a future where the physical truth of our built environment becomes as transparent and queryable as any digital database, creating a more equitable, efficient, and sustainable foundation for human habitation and commerce.

{\footnotesize

}

\end{document}